\PassOptionsToPackage{table}{xcolor}

\documentclass[conference]{IEEEtran}
\IEEEoverridecommandlockouts
\usepackage{nicefrac}
\usepackage{siunitx}
\usepackage{array,framed}
\usepackage{booktabs}

\usepackage{
  float,
  epsfig,
  wrapfig,
  subcaption
}
\usepackage{textcomp, amssymb}
\usepackage{setspace}
\usepackage{latexsym,fancyhdr,url}
\usepackage{enumerate}

\usepackage{graphics}
\usepackage{xparse} % argument parsing -- \edist
\usepackage{xspace}
\usepackage{multirow}
\usepackage{csvsimple}
\usepackage{balance}
\usepackage{tcolorbox}
\usepackage{tabularx} % Make sure this is in your preamble

\usepackage{algorithm}
\usepackage{graphicx}
\usepackage{subcaption}
\usepackage{algpseudocode}

\def\BibTeX{{\rm B\kern-.05em{\sc i\kern-.025em b}\kern-.08em
    T\kern-.1667em\lower.7ex\hbox{E}\kern-.125emX}}

\title{ASRJam: Human-Friendly AI Speech Jamming to Prevent Automated Phone Scams}
\author{
Freddie Grabovski \\ Ben Gurion University of the Negev
\and
Gilad Gressel \\ Amrita Vishwa Vidyapeetham
\and
Yisroel Mirsky \\ Ben Gurion University of the Negev
}

\begin{document}

\maketitle

\begin{abstract}
Large Language Models (LLMs), combined with Text-to-Speech (TTS) and Automatic Speech Recognition (ASR), are increasingly used to automate voice phishing (vishing) scams. These systems are scalable and convincing, posing a significant security threat. We identify the ASR transcription step as the most vulnerable link in the scam pipeline and introduce \textbf{ASRJam}, a proactive defence framework that injects adversarial perturbations into the victim's audio to disrupt the attacker’s ASR. This breaks the scam’s feedback loop without affecting human callers, who can still understand the conversation. While prior adversarial audio techniques are often unpleasant and impractical for real-time use, we also propose \textbf{EchoGuard}, a novel jammer that leverages natural distortions, such as reverberation and echo, that are disruptive to ASR but tolerable to humans. To evaluate EchoGuard’s effectiveness and usability, we conducted a 39-person user study comparing it with three state-of-the-art attacks. Results show that EchoGuard achieved the highest overall utility, offering the best combination of ASR disruption and human listening experience.
\end{abstract}


\section{Introduction}
\label{sec:intro}

Large Language Models (LLMs) are now widely used across many applications, demonstrating impressive progress in understanding and generating natural language \cite{openai2023gpt4, devlin2018bert, brown2020language}. When combined with text-to-speech (TTS) and automatic speech recognition (ASR) technologies, LLMs enable powerful new capabilities such as automated customer service, outbound sales, cold calling, and advanced virtual assistants. However, as these systems become more realistic and lifelike, they also raise significant security concerns.

% A significant emerging threat is the use of LLMs in automated technical support scams and voice phishing (vishing) attacks \cite{li2018close, gressel2024exploiting, figueiredo2024feasibility}. In these scenarios, attackers deploy LLM-powered agents enhanced with speech synthesis to place phone calls to victims. The goal is to deceive individuals—either by stealing money or by carrying out other forms of social engineering. These attacks are growing in scale and effectiveness, resulting in substantial financial losses and eroding public trust in voice-based communication systems \cite{cite1, cite2}.

LLMs have proven effective at generating phishing content that rivals human-written emails \cite{huang2024gpt4phish,li2024spearphishing}, contributing to a 703\% rise in credential phishing in 2024.\footnote{\url{https://www.infosecurity-magazine.com/opinions/resilience-age-ai-cybercrime/}} In parallel, deepfake voice scams have surged over 2,100\%\footnote{\url{https://sumsub.com/blog/deepfake-fraud-research-report/}}, driving \$12.5 billion in consumer losses in the U.S.\footnote{\url{https://www.infosecurity-magazine.com/opinions/resilience-age-ai-cybercrime/}} The integration of LLMs with speech synthesis into real-time, automated scam agents is the inevitable conclusion~\cite{gressel2024exploiting}. Recent studies show this pipeline is already practical: an academic study found an AI powered vishing system tricked 52\% of participants into revealing sensitive information \cite{figueiredo2024viking}; UIUC researchers demonstrated autonomous LLM agents executing scams like OTP theft with up to 60\% success\footnote{\url{https://www.bleepingcomputer.com/news/security/chatgpt-4o-can-be-used-for-autonomous-voice-based-scams/}}; and end-to-end vishing attacks using ChatGPT and voice cloning were successfully deployed in academic trials \cite{toapanta2024aidriven}.

\begin{figure}[t]
\centering
    \includegraphics[width=0.99\columnwidth]{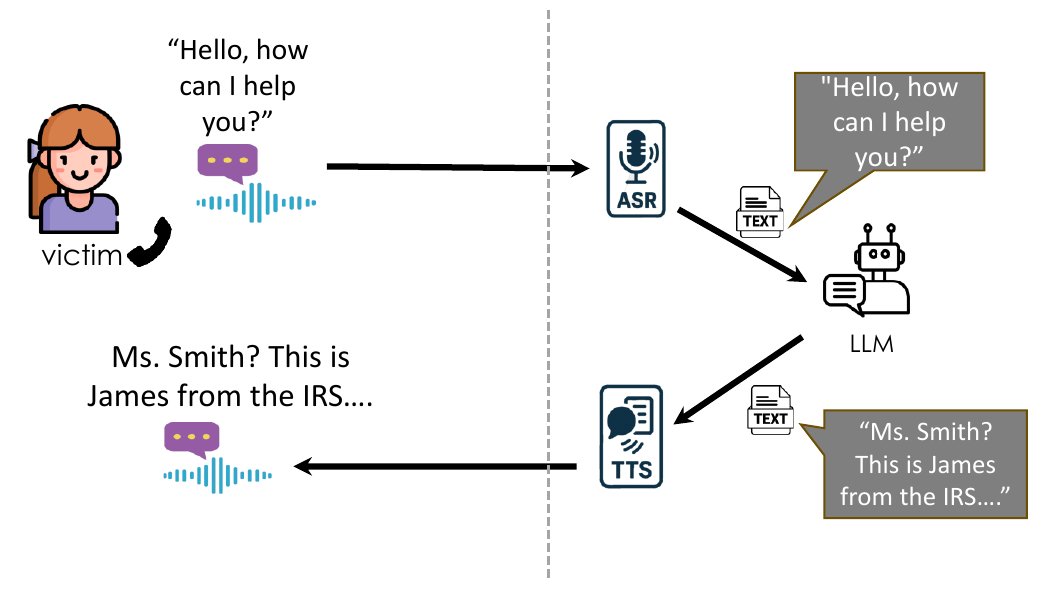}
    \caption{AI-driven scam call pipeline. The attacker uses automatic speech recognition (ASR) to transcribe the victim’s speech, a large language model (LLM) to craft context-aware responses, and text-to-speech (TTS) synthesis to generate a convincing, interactive scam.}
\label{fig:scam-pipeline}
\end{figure}

Voice agents operate by chaining together a sequence of neural networks to handle calls in real time: (1) ASR transcribes the victim’s speech into text, (2) An LLM generates an appropriate textual response, and (3) TTS synthesizes that response into natural-sounding audio. This pipeline enables scalable voice interactions that can convincingly impersonate trusted entities and extract sensitive information from victims as seen in Figure~\ref{fig:scam-pipeline}.

In order to accomplish attackers need not be technically savvy as there are already publicly available services that implement this full stack~\footnote{https://vapi.ai/, https://synthflow.ai/, https://www.bland.ai/, https://poly.ai/}. Moreover, the safeguards on these platforms are weak; it is often trivial to bypass LLM safeguards designed to prevent malicious use \cite{gressel2024exploiting,yu2024don,takemoto2024all}.

Traditional defences, such as passive detection, are failing as synthetic voices become indistinguishable from real ones. These methods rely on artifacts and imperfections that are quickly disappearing with advancing AI generation techniques ~\cite{muller2022does, zhang2025audio}. \footnote{https://uwaterloo.ca/news/media/how-secure-are-voice-authentication-systems-really} To stay ahead, we need \textbf{proactive} and \textbf{robust} defences that actively disrupt the attacker’s pipeline—targeting weaknesses AI cannot easily defend against.

In this paper we propose two novel contributions to this effort:

% proactive defence
% finding the weakest link - a very weak link is the ASR component - if the LLM cannot correctly "hear" the victim then their response will become incoherent and the victim will be protect.
% in order to disrupt the ASR we realize that we can use adversarial attacks (cite cite) to defend. To this end we propose a Jamming framework. That is we can inject some perturbation into the speech that breaks the ASR. Our framework allows for any user to defend against these attacks by insert an ASR attack into their phone conversations. However, in order for this defence to be effective a few criteira must be met. ... .

% practical ASR jammer
% The ASR jammer neends to be: universal - it should work against all ASR models with no prior information required, zero shot (no learning needed, ) it has to work immediately, it should not disrupt the listening experience for a human! Only for an adversary. Therefore it should be pleasant sounding (it will be audible, bbut it won't occlude the human from understand the speech, only disrupts the adversary ASR). It shoudl be effective in a short peirod of time so that it can be turned off once the conversation is secured.

\paragraph{Proactive defence}  
We identify the automatic speech recognition (ASR) component as the weakest link in the attacker’s pipeline. If the ASR fails to correctly transcribe the victim's speech, the downstream language model (LLM) produces incoherent or irrelevant responses, effectively neutralizing the scam. Our key insight is that by disrupting ASR performance, we can break the attack chain. To this end, we propose a proactive defence framework based on universal adversarial perturbations, carefully crafted noise added to the audio signal that confuses ASR systems while leaving human comprehension intact~\cite{ carlini2018audio, alzantot2018did, neekhara2019universal}.

This defence framework, which we call \textbf{ASRjam}, can be deployed by the user during phone calls to disrupt potential LLM-driven scams. To our knowledge, this is the first work to propose the use of ASR jamming as a proactive defence against automated voice-based social engineering attacks. ASRjam operates under a strict black-box, zero-query assumption: it has no knowledge of the ASR model’s architecture or parameters and cannot rely on querying the model during runtime~\cite{abdullah2021hear, ge2023advddos, wu2023kenku}. Despite these constraints, we demonstrate that effective, real-time ASR disruption is both feasible and practical.

\paragraph{Practical ASR Jamming}  
A major challenge with adversarial audio perturbations is that they tend to produce noise that is highly obtrusive to human listeners, rendering them impractical for use in everyday conversations. In our scenario, however, the jammer may be active during all suspicious calls, meaning that even legitimate human callers could be exposed to this noise. If the defence degrades call quality for genuine users, it will likely be disabled, undermining its utility.

To address this, we propose a novel ASR jamming algorithm called \textbf{EchoGuard}. EchoGuard is designed to be: (i) \emph{zero-shot}, working across a wide range of ASR models without any prior information, (ii) \emph{universal}, functional with any given input, (iii) \emph{real-time}, efficient enough for live deployment, and critically, (iv) \emph{human-compatible}, audible but not obtrusive. EchoGuard achieves this by harnessing naturally occurring sound distortions that humans are evolutionarily and cognitively equipped to tolerate: echoes, reverberations, or subtle modulations. These distortions degrade ASR accuracy due to biases in model training and architectural limitations, but remain intelligible to human listeners who are far better adapted to noisy environments.

To demonstrate the effectiveness of EchoGuard, we evaluate it against three state-of-the-art (SOTA) adversarial attacks that meet our criteria: they are black-box (require no access to the ASR model), zero-shot (require no interaction or queries), and universal (work across inputs) \cite{wu2023kenku, ge2023advddos, abdullah2021hear}. We perform experiments across three benchmark datasets, TEDLIUM, SPGISpeech, and LibriSpeech \cite{panayotov2015librispeech, tedlium, librispeech} and five popular ASR models \cite{hannun2014deep, baevski2020wav2vec, Pantev2019VoskAS, radford2023robust, Ravanelli2021SpeechBrain}. For each setup, we compare the performance of EchoGuard to the SOTA attacks on three criteria: attack success rate, clarity, and pleasantness.

In order to evaluate clarity and pleasantness of the attacks we conduct a 39 person user study to assess the perceived audio quality under each jamming method. 
On average, across the evaluated ASR models and datasets, our method (EchoGuard) achieved an Jamming success rate 19.6\% \textit{higher} than the average baseline performance. In addition, user ratings indicate our method achieved an average pleasantness score of 2.53 (on a 1-5 scale), compared to an average of 1.96 for the SOTA baselines, while being rated similarly for clarity (average difference of -0.09 points).\footnote{Audio examples comparing the perceptual quality of EchoGuard and baseline methods are available at: \url{https://sites.google.com/view/impulse-response-asr-attack/home}} Furthermore, EchoGuard demonstrates superior overall practicality, achieving an average utility index—a metric combining listener comfort (pleasantness, clarity) and ASR disruption (detailed in Section \ref{sec:metrics})—that is 2.12 times higher than the average baseline.

%FREDDIE list out why the other attacks are no good
% kenku - 
% kenansville - 
% advsddors - 

% Our work stands apart by focusing on natural audio transformations that are universally applicable without requiring access to the target ASR model. In addition, our method is evaluated through rigorous experiments and a user study, making it the first approach to prioritize comprehensibility alongside technical effectiveness.

% Our contributions include the following:
% \begin{itemize} 
% \item We propose a proactive defence framework that utilizes ASR jamming as a tool against automated scams
% \item We propose a novel real-time, zero-query, and untargeted adversarial transformation method that disrupts ASR models while preserving the human listening experience. 
% \item To validate our defence framework and our novel EchoGuard algorithm, we perform the first known user study comparing the pleasantness of adversarial audio transformations, ensuring practical applicability in real-world scenarios. 
% \end{itemize}

% This paper provides a comprehensive evaluation of our method across diverse ASR models, highlighting its robustness, scalability, and alignment with real-world needs. By bridging the gap between adversarial research and practical defences, our approach contributes to the development of proactive measures against the growing threat of AI-driven scams.

Our contributions include the following:
\begin{itemize} 
\item We are the first to propose \textit{ASR jamming} as a proactive defence against automated voice scams. Our framework, \textbf{ASRjam}, targets the weakest link in the attacker’s pipeline, speech recognition, disrupting LLM-driven vishing attacks without affecting human intelligibility.
\item We introduce \textbf{EchoGuard}, a novel blackbox, zero-query, and universal adversarial transformation method that reliably disrupts ASR systems while maintaining a comfortable listening experience for humans.
\item We validate our defence framework through the first known user study focused on the \textit{pleasantness} of adversarial audio transformations, demonstrating EchoGuard’s practical deployability in real-world scenarios.
\end{itemize}

\section{Background and Related Work}

In this section, we provide an overview of key concepts in audio processing and adversarial attacks on ASR systems, establishing the technical context for our work.

% \subsection{Representation of Audio Signals}

% Audio signals are typically represented as continuous waveforms that describe pressure variations over time. Mathematically, an audio signal \( x(t) \) is a function of continuous time \( t \). In practical digital processing, the signal is sampled at regular intervals. A discrete-time audio signal is thus represented as:

% \[
% x[n], \quad n = 0, 1, 2, \dots, N-1
% \]

% where \( x[n] \) denotes the amplitude of the waveform at discrete time index \( n \), and \( N \) is the total number of samples.

% The sampling frequency (or sample rate) \( f_s \), measured in Hertz (Hz), defines how many samples per second are captured from the continuous signal. In this work, we use a sample rate of \( f_s = 16000\, \text{Hz} \).

\subsection{Automatic Speech Recognition (ASR)}

\label{sec:asr-background}

% Automatic Speech Recognition (ASR) systems are a critical component in enabling machines to transcribe spoken language into text. These systems have seen significant advancements in recent years, driven by deep learning techniques, and are now integrated into various applications, from voice assistants to automated customer service platforms.

Modern Automatic Speech Recognition (ASR) systems predominantly utilize deep learning architectures to transcribe spoken language into text. Central to these systems are neural networks such as Recurrent Neural Networks (RNNs), Convolutional Neural Networks (CNNs), and, more recently, Transformers. These models process audio features—like spectrograms or mel-frequency cepstral coefficients (MFCCs)—and learn to map them directly to textual representations~\cite{prabhavalkar2023end}.

Traditional ASR pipelines consisted of separate stages: acoustic modeling, language modeling, and decoding. However, contemporary approaches often unify these steps into end-to-end frameworks. Transformer-based architectures, in particular, have gained prominence due to their ability to model long-range dependencies and contextual relationships across time, leading to improved robustness and accuracy~\cite{kheddar2024automatic}.

ASR systems are evaluated using metrics like Word Error Rate (WER), which measures the accuracy of the transcription by calculating the number of word substitutions, insertions, and deletions relative to a reference text.

Despite their progress, ASR systems remain vulnerable to various challenges, including noisy environments, accents, and adversarial attacks. Adversarial attacks, in particular, exploit subtle weaknesses in ASR models to produce incorrect transcriptions, posing a security risk in applications that rely on accurate speech recognition.

\subsection{Adversarial Attacks on ASR Systems}\label{sec:adv_attacks_asr}

Adversarial attacks on automatic speech recognition (ASR) systems have been widely explored in both white-box and black-box settings. In white-box attacks, the adversary has full access to the ASR model’s parameters and architecture, enabling precise crafting of perturbations which when added to the audio will cause the ASR system to break~\cite{carlini2018audio, yuan2018commandersong, schonherr2018adversarial, neekhara2019universal}. In contrast, black-box attacks assume no internal knowledge of the model, relying instead on probing the system’s outputs to infer vulnerabilities \cite{abdullah2021hear, taori2019targeted, liu2024difattack}. These attacks can be either targeted, forcing the ASR to output a specific incorrect transcription or untargeted, where the goal is simply to reduce transcription accuracy.

% In this work, we focus on a practically important subset of adversarial attacks: those that are \textit{black-box}, \textit{universal}, and \textit{zero-query}. These properties are essential for an effective ASR jammer to be used as a defence. \textbf{Black box}: when a bot calls, we don't know which ASR is being used so our technique must not have assumptions on which model is used.  \textbf{Universal}: the jamming must be applied in real-time, so the most efficient way to do this is to apply the same perturbation universally to every audio frame sent to the caller. \textbf{Zero-query}: once a caller is connected we cannot practically send audio clips to the attacker to optimize the perturbation since the call has already started, therefore the perturbation cannot be optimized using query-based methods.   

In this work, we focus on a practically important class of adversarial attacks characterized by three critical properties: \textit{black-box}, \textit{universal}, and \textit{zero-query}. These constraints are essential for deploying an effective ASR jammer in real-world defence scenarios. \textbf{Black-box}: During scam calls, the defender has no knowledge of which ASR system the attacker is using. Therefore, the jammer must not rely on any assumptions about the model’s architecture or parameters. \textbf{Universal}: To operate in real time efficiently the jammer should apply the same perturbation to every outgoing audio frame. \textbf{Zero-query}: Once a call has started, it is infeasible to send trial audio clips to the attacker’s ASR to refine the perturbation. Thus, query-based optimization is impractical, and the jammer must operate without interaction with the adversarial model.

Several ASR jammers have been proposed recently that meet these requirements.

\textbf{AdvDDoS}~\cite{ge2023advddos} proposes a zero-query universal adversarial perturbation (UAP) attack targeting commercial speech recognition systems without the need for model queries or internal knowledge. It generates perturbations by extracting robust acoustic features from a surrogate model (e.g., DeepSpeech) and performing feature inversion to craft a universal noise pattern. The result is a reusable perturbation that consistently misleads diverse ASR models.

\textbf{Kenansville}~\cite{abdullah2021hear} introduces phoneme-level adversarial perturbations by selectively removing or suppressing low-amplitude spectral components using Discrete Fourier Transform (DFT). By discarding frequency bands with minimal energy, Kenansville induces significant transcription errors in ASR while preserving human intelligibility.

\textbf{Kenku}~\cite{wu2023kenku} crafts audio perturbations through dual-objective optimization in the Mel-frequency cepstral coefficient (MFCC) feature space, targeting minimal perceptibility and maximal misclassification towards a predefined command. Typically, Kenku balances perturbation strength against transcription accuracy via binary search optimization.\footnote{The original Kenku method operates offline. Our implementation modifies it to align with the real-time requirements of our evaluation.}

% \paragraph{Contrast to Prior Works}

% While we build on insights from these prior attacks, none were designed with human listeners in mind. Their perturbations, though often intelligible, are perceptually harsh and impractical for interactive scenarios. In contrast, we introduce \textbf{EchoGuard}, a universal, zero-query, and human-compatible jammer that operates in real time and preserves audio quality for legitimate callers.  

\paragraph{Contrast to Prior Works}

% % Adversarial perturbations are the perfect solution in our context because, by definition, the perturbation is not supposed to be obvious to a human observer.
% However, in contrast to white box generated adversarial perturbations, black box generated perturbations are typically much more obvious and disruptive to humans. This is common in black box attacks since the larger perturbations are necessary to overcome gradient alignment [].

While we build on insights from the prior work, we also introduce two novel concepts. First, we introduce ASR jamming as a proactive defence against neural network-driven scams, a use case not explored in past work, which has focused exclusively on ASR evasion, not protection. Adversarial perturbations are especially well-suited to be used in this defence because they are designed to be imperceptible to humans while remaining disruptive to machine learning models.

However, while adversarial perturbations are theoretically well-suited for defence, in practice black-box perturbations, unlike their white-box counterparts, tend to be more perceptible to human listeners. This is because black-box attacks lack access to model gradients, and thus require stronger, more noticeable perturbations to achieve the same effect. As a result, most existing black-box jammers introduce audible artifacts that limit their usability in real-time, human-facing settings~\cite{ambra, wang2024practicalsurveyemergingthreats}.

Therefore as our second contribution we design \textbf{EchoGuard}, the first ASR jammer built for real-world use. While prior black-box, zero-query jammers exist, none are suitable for interactive scenarios; their perturbations, though often intelligible, are perceptually harsh and impractical for interactive scenarios. EchoGuard fills this gap with a human-compatible, real-time jammer that preserves call quality while disrupting ASR.\footnote{Audio samples demonstrating the perceptual quality of EchoGuard compared to baseline methods are available at: \url{https://sites.google.com/view/impulse-response-asr-attack/home}}

% However, none of these works evaluate the human perceptibility or acceptability of the resulting audio, which is critical for defensive use in real-world conversations.

% We draw from these approaches to evaluate the performance of our proposed jammer, \textbf{ASRJam}, which is designed to be universal, zero-query, and human-compatible for proactive defence against scam calls.

\section{Threat Model}

In our threat model, the adversary seeks to automate large-scale scam and voice phishing attacks through an AI-driven conversational pipeline. This can be accomplished though existing technologies and services that provide convenient, modular building blocks for automatic speech recognition (ASR), large language models (LLMs), and text-to-speech (TTS).\footnote{https://vapi.ai/, https://synthflow.ai/, https://www.bland.ai/, https://poly.ai/} By chaining these services together, these platforms enable attackers to conduct real-time, interactive conversations with potential victims, with minimal human intervention.

The primary motivation for attackers to adopt such a pipeline is scalability and improved attack success rates: First, by employing advanced TTS systems, an attacker can customize the voice output, selecting accents or authoritative tones to appear more convincing and culturally aligned with the target. Second, continuous improvements in LLMs allow for highly coherent and context-aware dialogue, enabling the scam to scale more effectively—covering a larger pool of potential victims—while maintaining personalized and believable interactions.

We assume that the attacker must use models and libraries that can run in real-time to maintain the illusion of a real human talking on the other side of the phone. In addition, because the adversary is aiming for scalability, it is not possible to 'hot-seat' and swap in human into a call if the ASR starts to fail.

\section{Defence Framework}
% give overview of defence framework (high level idea/objective) like in the intro. Note that can be on the users phone since some are realime [][]. point to figure that shows the pertubation being added to call to protect user and that it does nto both human caller too. 

% Our defence framework, ASRJam, is designed to proactively disrupt AI-driven scam calls by targeting their reliance on automatic speech recognition (ASR). The core idea is simple: inject carefully crafted perturbations into the user’s audio stream that degrade ASR performance while remaining perceptually unobtrusive to human callers. Since the scam pipeline depends on accurate transcriptions to generate convincing responses, even small ASR errors can cause the system to fail, exposing the scam. Crucially, this defence can run in real time on the user’s device, such as a smartphone, without needing external infrastructure or model access. Figure~\ref{fig:Attack_defence_Framework} illustrates how ASRJam adds perturbations during live calls, disrupting automated adversaries while preserving natural communication with legitimate human callers.

\begin{figure}[t]
\centering
    \includegraphics[width=0.99\columnwidth]{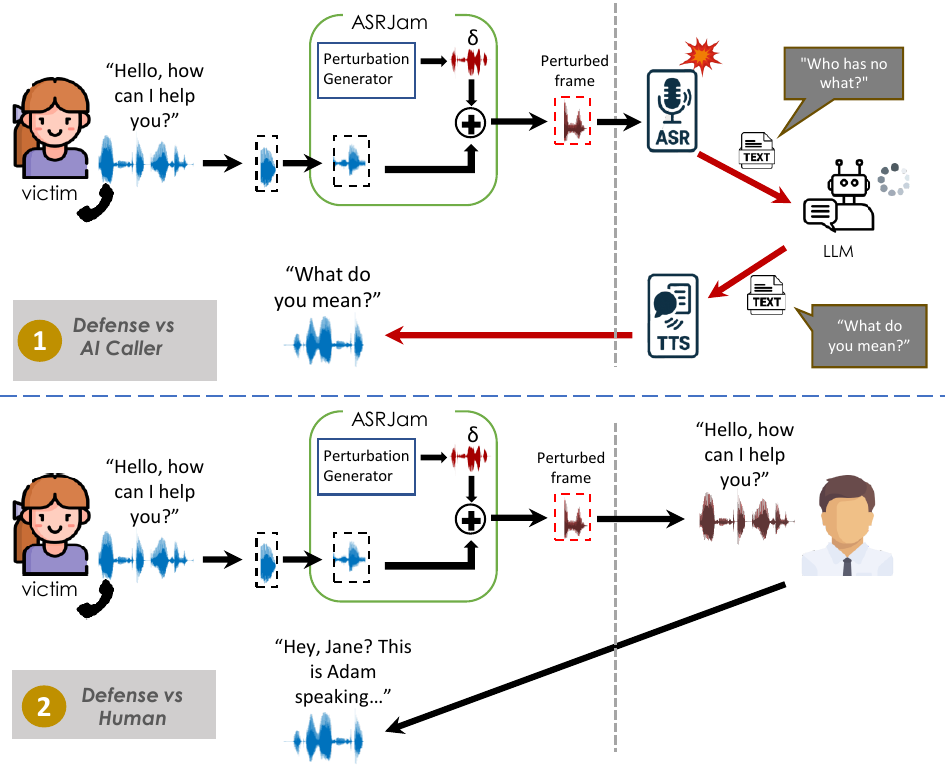}
    \caption{Illustration of the ASRJam defence framework. (1) When the incoming call originates from an AI-powered caller, subtle acoustic perturbations introduced by ASRJam degrade the adversary's speech recognition accuracy, disrupting the automated scam pipeline (ASR → LLM → TTS). (2) When the caller is human, the acoustic perturbations have minimal perceptual impact, ensuring effective communication.}
\label{fig:Attack_defence_Framework}
\end{figure}

ASRJam is a proactive defence framework designed to break the conversational loop that enables AI-driven scam calls. These attacks rely on automatic speech recognition (ASR) to transcribe a victim’s speech, which is then processed by a large language model (LLM) to generate real-time, persuasive responses. ASRJam disrupts this pipeline by injecting subtle, adversarial perturbations into the user’s audio during live calls, degrading ASR accuracy while remaining unobtrusive to human listeners. Unlike prior work that treats ASR jamming as an offensive tool, our approach repurposes it as a defensive mechanism against social engineering. Crucially, ASRJam runs in real time on end-user devices, requires no access to the attacker’s models, and introduces minimal perceptual overhead, making it practical for everyday use. Figure~\ref{fig:Attack_defence_Framework} illustrates how ASRJam selectively interferes with scam agents while preserving natural communication with human callers.

% Explain how the ASR jammer is preprared and applied: outgoing audio is split into frmaes. then each audio frame (eg 10ms) we genreate or apply a pre-prepaired pertubation to the frame. we can use any BB ASR attack. for example, if its X [] then first trian the BB model on surrogates.

\paragraph{Perturbation Generation and Application}

ASRJam operates by modifying the user’s outgoing audio stream in real time. The audio is first divided into short frames (e.g., 10 milliseconds), and a perturbation is drawn from a precomputed universal noise pattern. Each frame is then perturbed before being transmitted, ensuring continuous ASR degradation throughout the conversation. The framework is flexible: it supports any black-box, zero-query, universal adversarial attack, including both real-time generation methods and pre-trained perturbations.

% Explain that not all calls need to be jammed: can use contact or other white list to minimize interference with legitimate human calls. Can also be turned on by vticim manually when he/she does is suspicous. Doewsnt need to be intirecall, can be activated by vicitm when supiocis too.

\paragraph{Deployment and Usability}
To minimize interference with legitimate conversations, ASRJam does not need to be active during every call. Users can maintain a whitelist of trusted contacts to only activate the jammer on unknown callers. Alternatively, the system can be configured to activate only when the user suspects something is off, either through a manual trigger or heuristics based on call behaviour. Importantly, ASRJam can be engaged dynamically; it does not need to run for the entire duration of a call. This flexibility ensures that the defence remains practical and user-friendly, while still providing robust protection when a scam attempt is suspected.

% limitaitons: the defender in this case is limited to theuse of zero-query black box attacks (such as [[][]][[]). The ASR has to be realtime -=e.g. [] is BB but not not raltime bc you must find X.

\paragraph{Limitations}
A key constraint of our defence framework is the requirement to operate under a zero-query, universal, black-box threat model. Since the defender has no access to the adversary’s ASR system and cannot query it during runtime, only attacks that generalize across models and inputs can be used. This excludes many high-precision adversarial methods that rely on white-box access or per-sample optimization. For example, while \cite{chen2020devil} qualifies as a black-box attack, it is not usable in our setting because it requires query-based tuning for each input, making it incompatible with real-time deployment. Therefore, ASRJam is limited to universal, zero-query perturbations that strike a balance between effectiveness and practical deployability.

While prior zero-query, universal attacks like AdvDDoS \cite{ge2023advddos} and Kenansville \cite{abdullah2021hear} satisfy the technical constraints of our defence setting, they neglect a critical factor for real-world deployment: human usability. Their perturbations are often harsh, distracting, and unpleasant to listen to, making them unsuitable for interactive phone conversations. To address this gap, we introduce EchoGuard, a novel ASR jamming algorithm specifically designed for our defence framework. EchoGuard is a black-box, zero-query, and universal adversarial perturbation that is designed with natural audio distortions, such as echoes, that are cognitively tolerable for humans but disruptive to ASR systems.

\section{EchoGuard}

% \subsection{Impulse Response and Convolution}

% A room impulse response (RIR), which is analogous to how a room responds to an echo. characterizes how an acoustic environment responds to a short, instantaneous sound burst (an impulse). Formally, an impulse response is denoted as \( h[n] \), and it fully describes the acoustic properties, such as echo and reverberation, of a given environment.

% Given an original audio signal \( x[n] \), the reverberated audio \( y[n] \) resulting from passing \( x[n] \) through an acoustic environment characterized by impulse response \( h[n] \) is obtained by discrete convolution:

% \[
% y[n] = (x * h)[n] = \sum_{k=-\infty}^{\infty} x[k]\,h[n - k]
% \]

% In practice, signals have finite length, simplifying to:

% \[
% y[n] = \sum_{k=0}^{L-1} x[k]\,h[n - k], \quad n = 0, 1, \dots, N+L-2
% \]

% where \( L \) is the length of the impulse response.

\subsection{The Inspiration for EchoGuard}

\captionsetup[subfigure]{skip=5pt} % adds space between image and subcaption

\begin{figure}
    \centering

    \begin{subfigure}{\linewidth}
        \centering
        \includegraphics[width=0.9\linewidth]{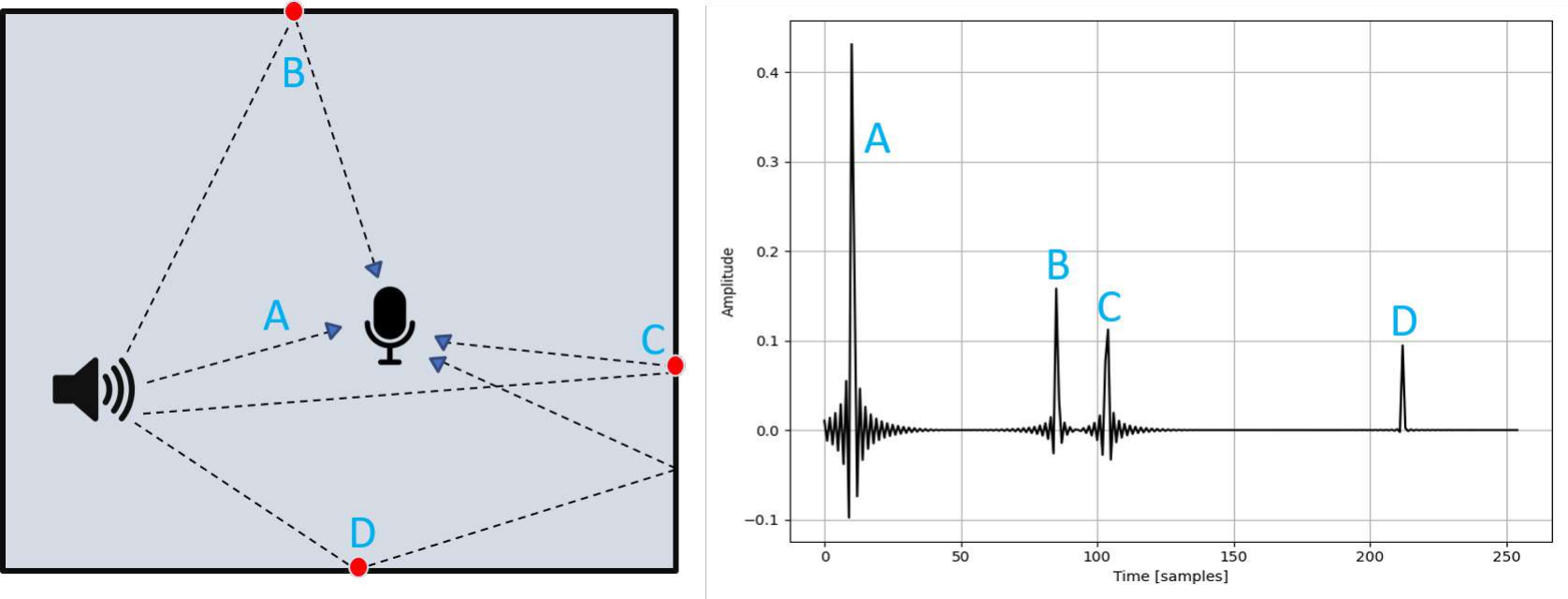}
        \caption{Reverberation effects due to room geometry.}
        \label{fig:reverb}
    \end{subfigure}

    \vspace{1em}

    \begin{subfigure}{\linewidth}
        \centering
        \includegraphics[width=0.9\linewidth]{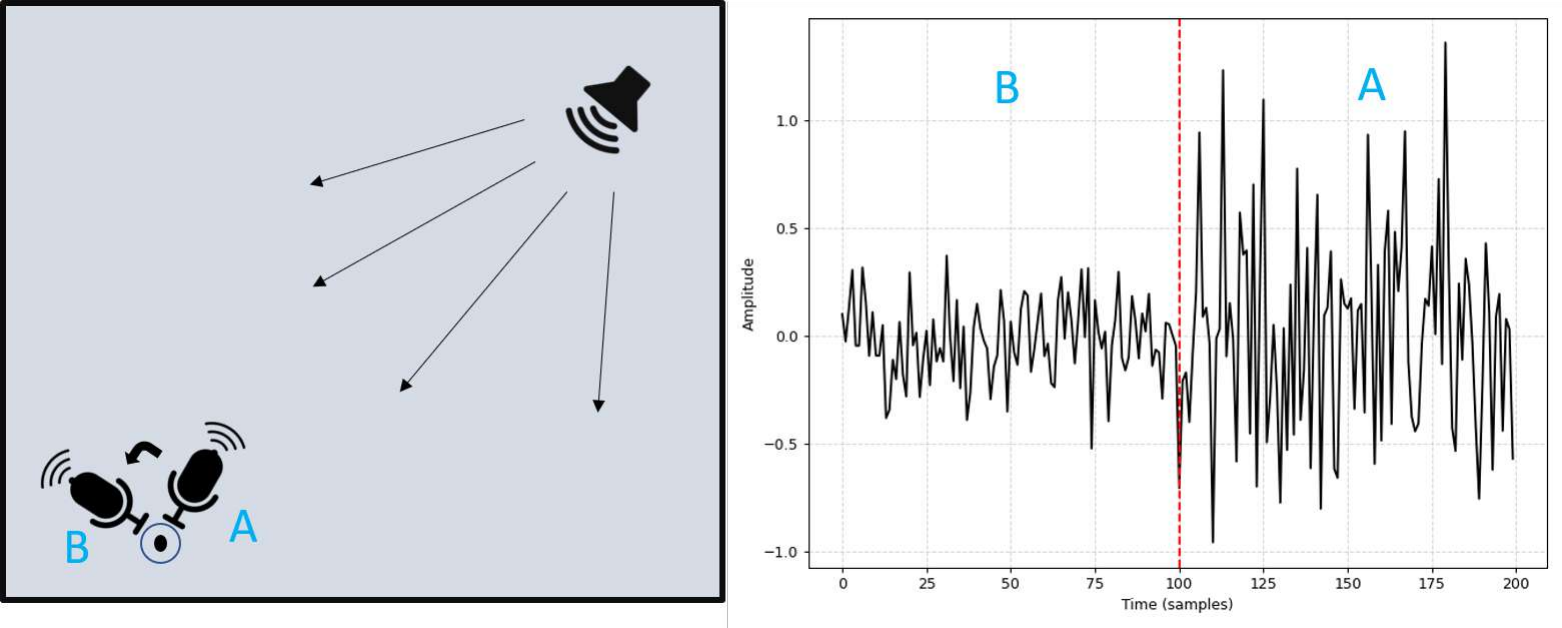}
        \caption{Directional distortion caused by microphone oscillation.}
        \label{fig:direction}
    \end{subfigure}

    \vspace{1em}

    \begin{subfigure}{\linewidth}
        \centering
        \includegraphics[width=0.9\linewidth]{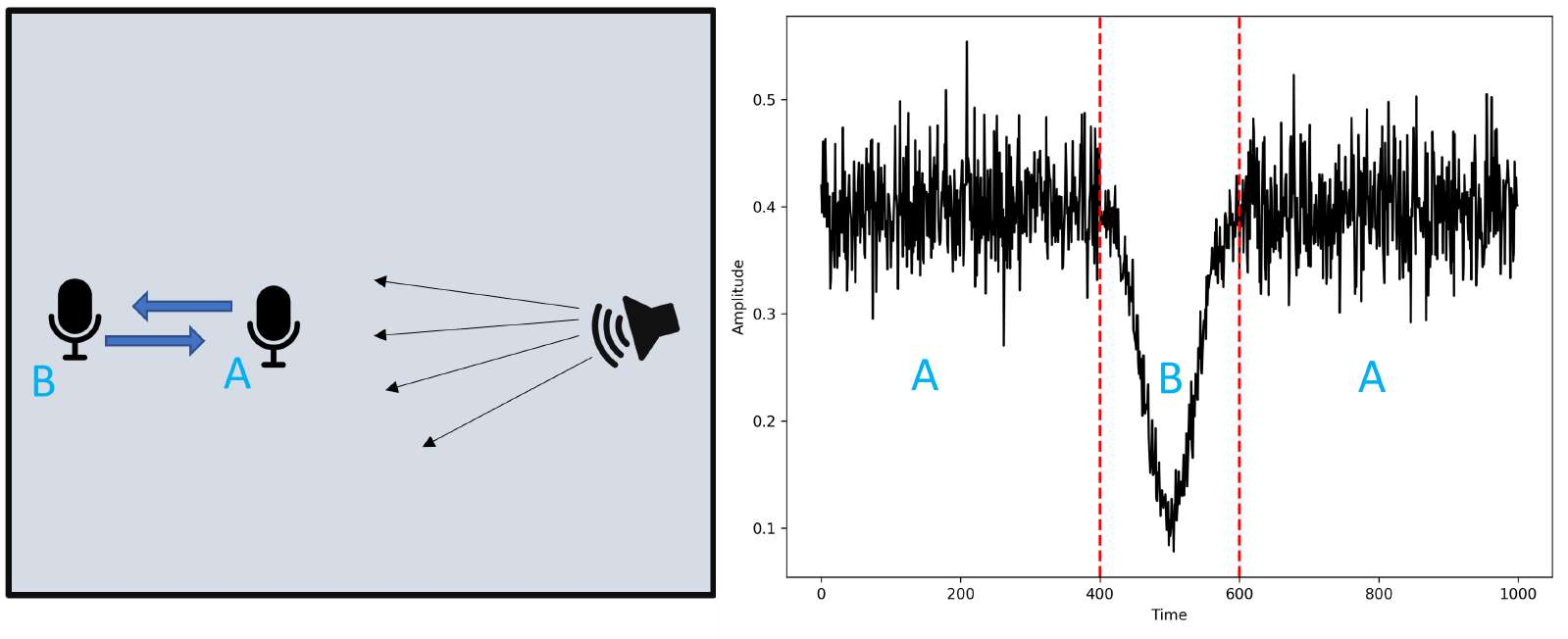}
        \caption{Transient attenuation during short intervals.}
        \label{fig:attenuation}
    \end{subfigure}
    
    \vspace{1.2em}

    \caption{Illustration of the three natural acoustic transformations used to develop \textsc{EchoGuard}.}
    \label{fig:insights}
\end{figure}

In designing EchoGuard, our goal was to create black box universal perturbations that are easy for humans to tolerate yet difficult for ASR systems to manage. This led us to focus on natural sounds, distortions the human auditory system is evolutionarily equipped to handle. In natural listening, speech signals are shaped by various everyday acoustic transformations, including echoes in reverberant environments, moment-to-moment direction changes of a talker, and brief modulations in volume. Human auditory perception is remarkably resilient under these conditions, leveraging neural adaptation, streaming, and contextual cues to preserve intelligibility~\cite{tsironis2024adaptation,luo2024cortical,garcia2024sensory,gao2024original}. 

Modern ASR pipelines, however, lack such sophisticated mechanisms and are prone to error when confronted with these naturally occurring distortions. In particular echoes and reverberations are largely absent from ASR training datasets. This is because the vast variability of real-world echoes would require an enormous and complex dataset of Room Impulse Responses (RIRs), risking overfitting and diminishing generalizability. This gap in robustness between humans and ASR forms the basis of \textsc{EchoGuard}, which deliberately applies \textit{three} classes of natural transformations to compromise ASR without substantially affecting human comprehension.

We will now review the three transformations that we used to create \textsc{EchoGuard}

\subsubsection{Reverberation}

In everyday listening, speech signals are naturally modified by reverberation, the result of sound reflections from walls, ceilings, and other surfaces in an enclosed space. Reflections from surrounding surfaces can blur temporal details crucial for machine transcription. Reverberation fills the valleys of the speech envelope, sometimes eliminating essential modulations~\cite{luo2024cortical,gao2024original}, while also introducing longer-term echoes that degrade spectral-temporal landmarks~\cite{tsironis2024adaptation}. In simple terms, reverberation smears out the sound energy over time and frequency, making it harder for automatic speech recognition (ASR) systems to distinguish between individual speech sounds.

Although the human brain can dynamically segregate the primary voice from its echoes and partially reconstruct missing rhythmic cues~\cite{gao2024original}, conventional ASR systems trained on stationary patterns struggle to compensate for reverberant smearing. By manipulating room impulse responses and absorption, we systematically produce these detrimental overlaps (Fig.~\ref{fig:reverb}) to degrade ASR while minimally impacting human listeners, whose auditory systems are evolutionarily adapted to handle reverberant environments.

A key concept in understanding reverberation is the \emph{impulse response}, which characterizes how an acoustic environment reacts to a brief, ideally instantaneous, sound burst. In physical terms, the impulse response encapsulates all the reflection delays and attenuations that a sound undergoes in a given space.

Mathematically, if we denote the original speech signal as \( x(t) \) and the impulse response of the room as \( h(t) \), then the reverberated signal \( y(t) \) is obtained by the convolution:
\[
y(t) = \bigl(x * h\bigr)(t) = \int_{-\infty}^{\infty} x(\tau)\,h(t-\tau)\,d\tau.
\]
Here, \( h(t) \) is typically modeled as a sum of delayed impulses:
\[
h(t) = \sum_{i=0}^{N-1} a_i\,\delta(t-\tau_i),
\]
where \( a_i \) are the attenuation coefficients, \( \tau_i \) are the corresponding time delays for each reflection, and \( \delta(t) \) is the Dirac delta function. This formulation effectively represents the environment as a series of time-shifted and scaled replicas of the original impulse (see Fig.~\ref{fig:reverb}).

The physical characteristics of reverberation in a room are determined by several \emph{acoustic scene parameters}. We focus on two critical parameters:

\begin{figure}
\centering
\includegraphics[width=0.95\columnwidth]{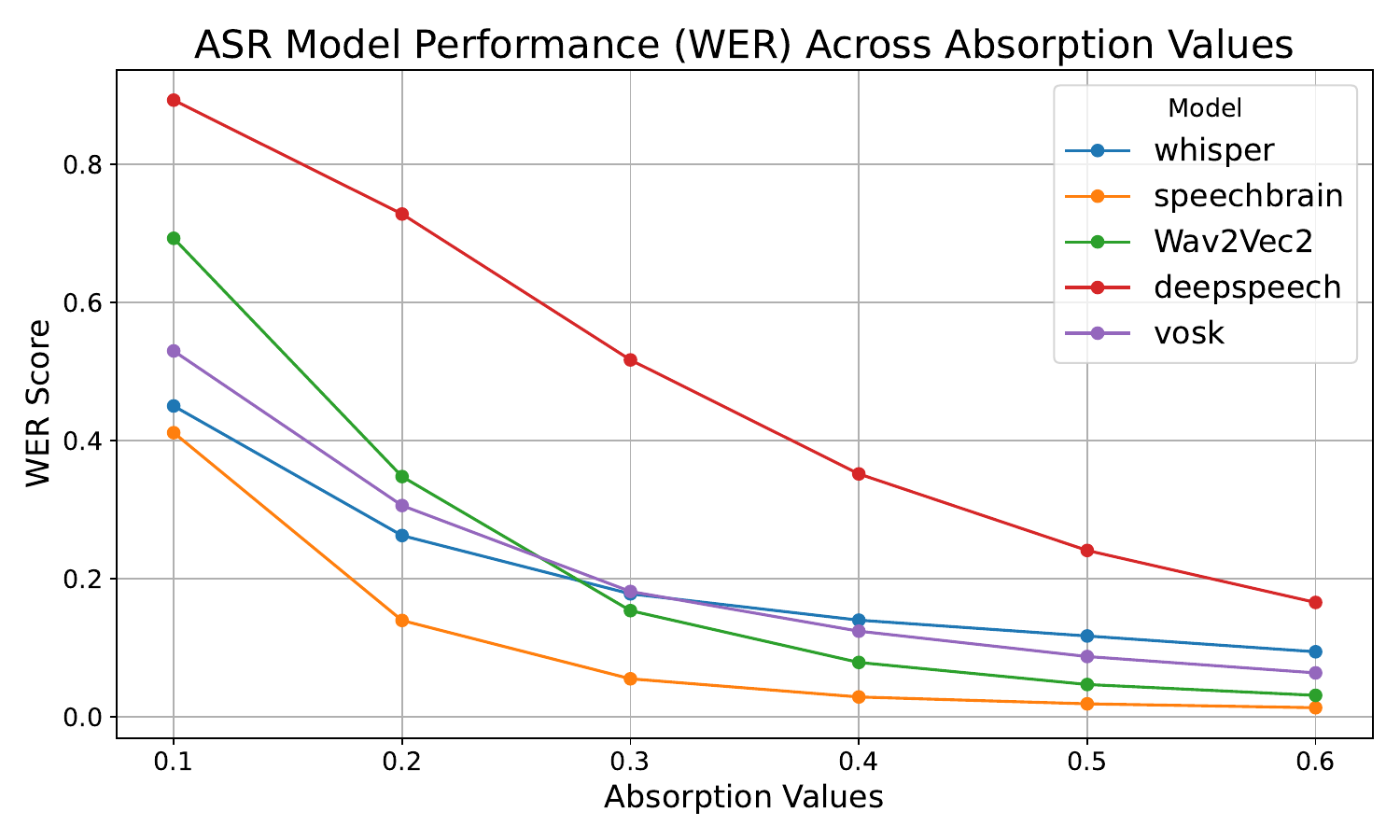}
\caption{Word Error Rates (WER) performance (lower is better) of ASR models across increasing absorption coefficients.}
\label{fig:absorption_plot}
\end{figure}

\begin{figure}
\centering
\includegraphics[width=0.95\columnwidth]{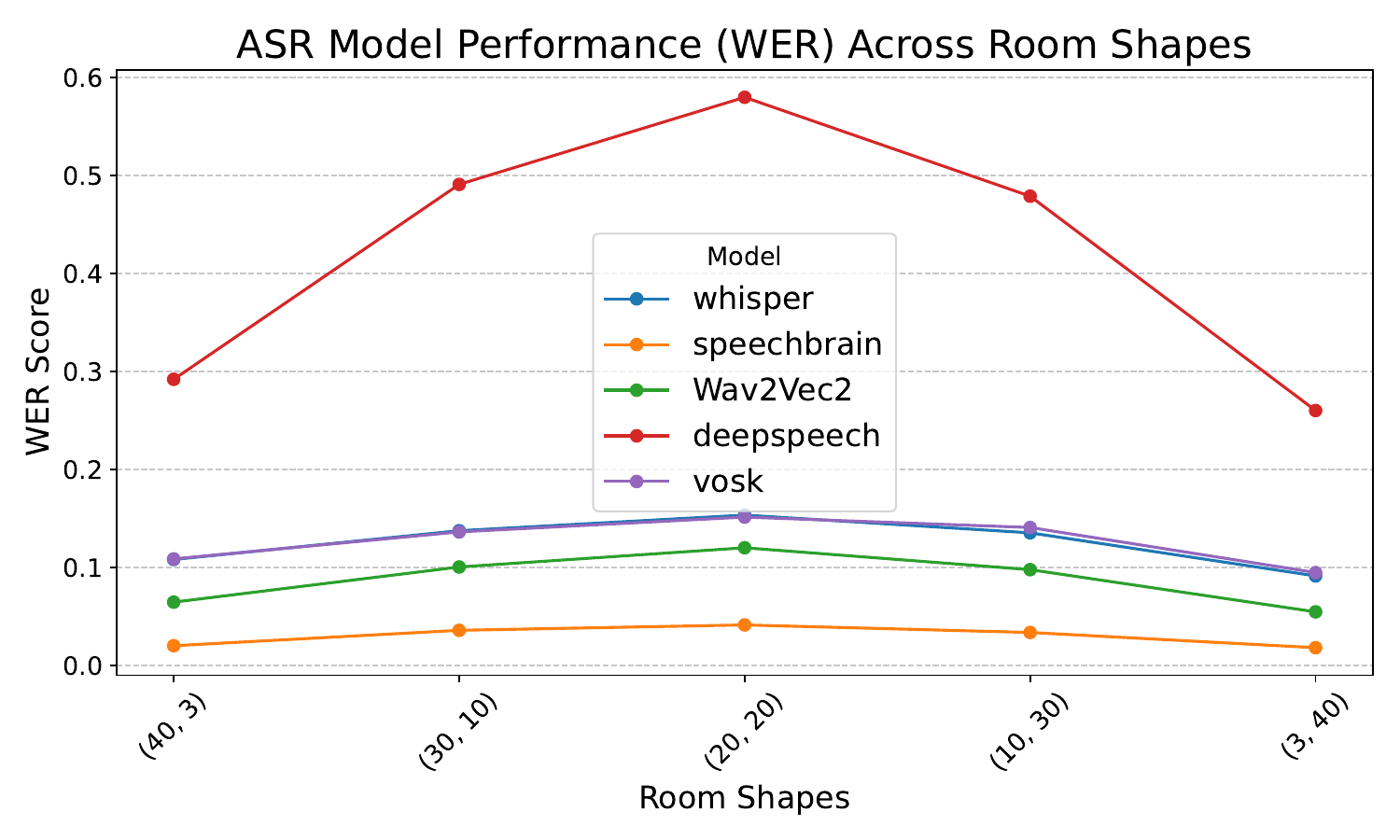}
\caption{Word Error Rate (WER) of different ASR models as a function of the normalized room aspect ratio, calculated as Length / (Length + Width). A ratio close to 0 or 1 indicates a highly elongated room, while a ratio of 0.5 corresponds to a square room. Performance generally degrades as the room shape deviates significantly from square (ratio approaches 0 or 1).}
\label{fig:room_shape_plot}
\end{figure}

\begin{figure}
\centering
\includegraphics[width=0.95\columnwidth]{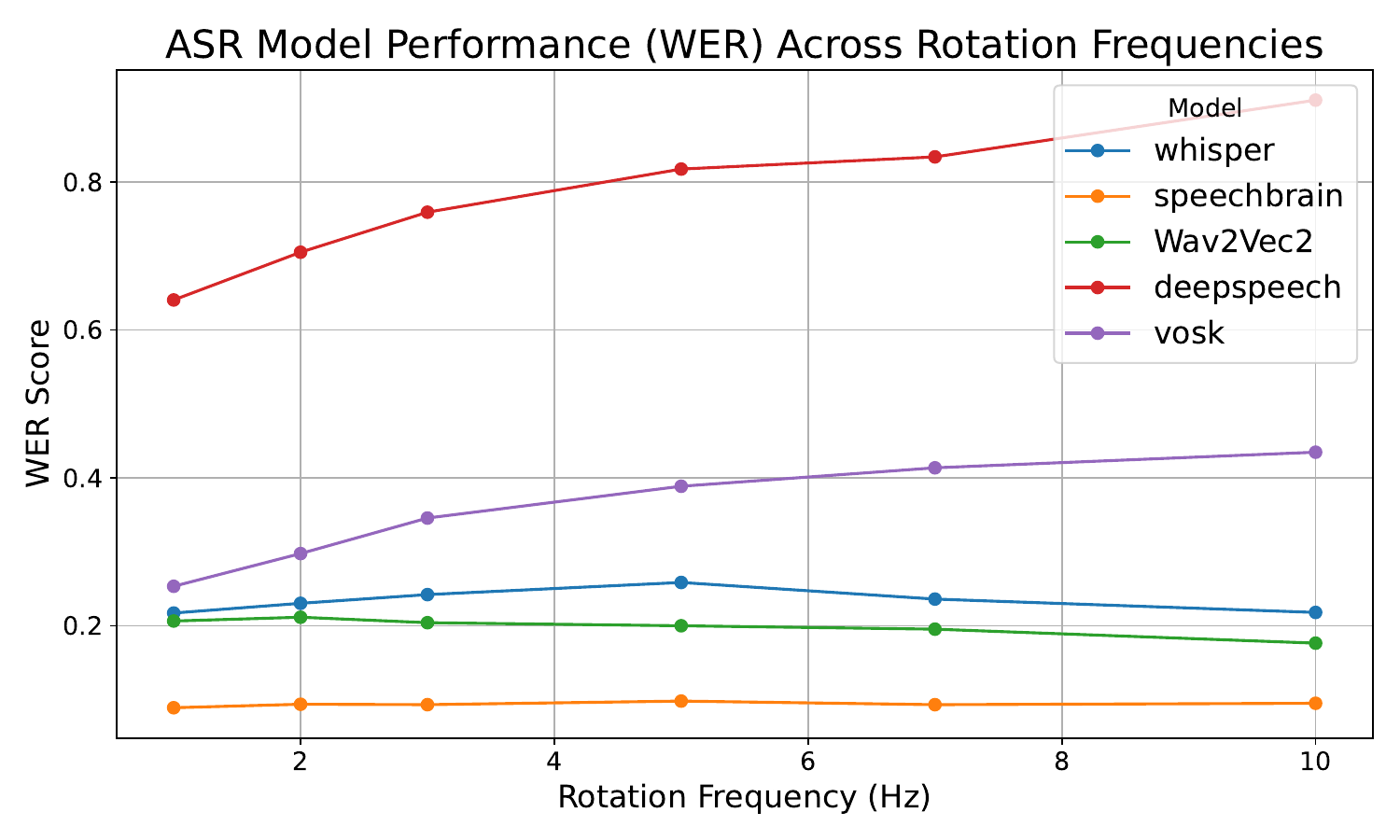}
\caption{Word Error Rates (WER) performance (lower is better) of ASR models across rotation frequencies.}
\label{fig:Directional_performance}
\end{figure}

\begin{itemize}
    \item \textbf{Surface Absorption:} This parameter quantifies the fraction of sound energy absorbed by the room surfaces. For example, highly absorptive materials (e.g., stone) reduce the intensity and duration of reflections, resulting in a weaker reverberant tail. In contrast, low absorption leads to high reverberation, characterized by prolonged echoes and overlapping acoustic signals. Such environments create challenging acoustic conditions for ASR, as the repeated and overlapping speech signals distort essential features like phoneme boundaries and spectral clarity, resulting in elevated transcription errors. As illustrated in Figure~\ref{fig:absorption_plot}, lower surface absorption values, corresponding to increased reverberation, generally result in higher transcription error rates for the evaluated ASR models.
    
    \item \textbf{Room Geometry:} The dimensions and geometry of a room influence the timing and distribution of sound reflections, shaping the reverberant field experienced by both humans and machines. We experimentally evaluate ASR robustness across rooms of varying dimensions while holding constant the speaker-microphone distance and surface absorption characteristics (Fig.~\ref{fig:room_shape_plot}). Results reveal that a $20\times20$~m room produces the highest ASR error, indicating that even regular-shaped spaces can generate reverberant conditions that significantly impair machine transcription.
\end{itemize}

\begin{tcolorbox}[colback=gray!5!white,colframe=black!75!black]
\textbf{Hypothesis:} Human auditory systems excel at disentangling the multiple reflections inherent in reverberation through sophisticated neural processing. Conversely, automatic speech recognition (ASR)
systems predominantly rely on clean acoustic patterns, making
them vulnerable to distortions introduced by reverberation. 
\end{tcolorbox}

\subsubsection{Motion-Induced Acoustic Variation (Microphone Oscillation)}

In natural conversational settings, speakers often change their head orientation and position, dynamically altering the directionality and intensity of their speech signals. Humans can effortlessly comprehend speech even when the speaker is moving their head or turning, due to robust directional hearing capabilities, which leverage differences in sound intensity and timing between ears. Specifically, human auditory perception integrates subtle and continuous shifts in the directionality of sound sources, allowing consistent speech comprehension despite dynamic auditory scenes~\cite{gao2024original}.

To simulate and exploit this phenomenon artificially, we introduce a dynamic acoustic variation in the form of a spinning directional microphone, modeled through varying directional impulse responses (IRs). Practically, this is implemented by cycling through precomputed impulse responses captured from microphones oriented at different angles relative to the speaker (Fig.~\ref{fig:direction}). The microphone array simulates motion at a defined \emph{rotation frequency} (measured in Hz), where a frequency of $1$ Hz corresponds to a complete rotation cycle occurring over one second. 

This setup mimics the auditory experience of a listener physically rotating their head while listening to speech, causing continuous changes in how the speech signal arrives at the listener’s ears. Humans easily compensate for this dynamic variation by naturally processing acoustic cues, preserving speech intelligibility. In contrast, ASR systems, optimized primarily for stationary acoustic patterns, struggle to reliably recognize speech when directional acoustic cues dynamically shift.

Experimentally, we validate this by systematically altering microphone orientation at various rotation frequencies and observing the resulting drop in ASR accuracy. Figure~\ref{fig:direction} shows that ASRs are disrupted by even modest rotational frequencies.

\begin{tcolorbox}[colback=gray!5!white,colframe=black!75!black]
\textbf{Hypothesis:} Humans naturally compensate for motion-induced acoustic variations due to their ability to leverage directional acoustic cues from stereo hearing. In contrast, ASR models, which depend on stable spectral-temporal patterns, struggle under dynamic directional variations introduced through simulated microphone rotations.
\end{tcolorbox}

\subsubsection{Short-term Acoustic Attenuation}

During everyday speech communication, brief fluctuations in audio intensity occur naturally. These transient attenuations, short periods during which speech volume momentarily decreases, can result from speaker movements, environmental noise, or temporary disruptions. Despite these short-duration interruptions, human listeners typically experience minimal or no loss in speech intelligibility. This robustness stems from the brain's remarkable ability to utilize contextual and linguistic cues to interpolate and reconstruct briefly attenuated or obscured acoustic segments~\cite{gao2024original}.

In contrast, automatic speech recognition (ASR) systems rely heavily on stable and continuous acoustic patterns. Even brief transient attenuation introduces substantial variability in acoustic features, such as spectral energy and formant stability, upon which ASR algorithms depend for accurate phoneme classification and subsequent transcription. Empirically, we observe that transient acoustic attenuations, occurring at high-energy speech points, significantly impair ASR system performance, causing degradation in transcription accuracy. Conversely, these brief attenuations result in very little perceptual degradation for human listeners, whose auditory processing easily compensates for such momentary intensity reductions.

\begin{tcolorbox}[colback=gray!5!white,colframe=black!75!black]
\textbf{Hypothesis:} 
Transient acoustic attenuation introduces minimal perceptual impact on human comprehension due to our cognitive and contextual reconstruction capabilities. Conversely, ASR systems experience performance degradation from these brief audio interruptions, providing an effective yet subtle mechanism for adversarial speech modification.
\end{tcolorbox}

\subsection{Methodology}

% Utilizing the insights gained from analyzing acoustic phenomena, we develop \textit{EchoGuard}, an adversarial audio transformer designed to dynamically modify incoming audio signals in real-time. The goal of \textit{EchoGuard} is to significantly degrade automated speech recognition (ASR) performance while maintaining clear intelligibility for human listeners.

% \subsubsection{Overview}

% At its core, \textit{EchoGuard} employs optimized acoustic transformations - specifically room acoustic configurations and microphone oscillations - to take advantage of differences in auditory processing capabilities between humans and ASR models. To systematically identify the optimal acoustic environment, we utilize an evolutionary algorithm (EA), a powerful optimization technique that mimics biological evolution.

Utilizing the insights gained from analysing acoustic phenomena, we develop \textit{EchoGuard}, an adversarial audio transformer designed to dynamically modify incoming audio signals in real-time. In order to create an acoustic transformation with EchoGuard we perform the following steps

% \subsubsection{EchoGuard Overview}

\begin{itemize}
    \item \textbf{Reverberation Profiles}: EchoGuard employs a genetic algorithm to precompute room parameters (e.g., dimensions, microphone/source positions, absorption coefficients) that yields an optimal reverberation profile which degrades ASR accuracy while preserving human intelligibility.
    
    \item \textbf{Microphone Oscillation}: EchoGuard applies the reverberations using the optimized room configurations and simulates directional microphone motion by blending multiple directional responses dynamically, introducing natural acoustic variation.
    
    \item \textbf{Transient Acoustic Attenuation}: EchoGuard intermittently attenuates short segments of audio using Gaussian-shaped modulation, introducing subtle disruptions that impact ASR systems with minimal perceptual cost to human listeners.
\end{itemize}

\subsubsection{Reverberation Profiles} % what is the transformation here? 

The first step in creating the adversarial perturbation with EchoGuard is to select an appropriate reverberation profile. We apply an evolutionary algorithm (EA) to identify optimal acoustic configurations that degrade ASR performance while preserving human intelligibility. EAs are stochastic optimization methods inspired by natural selection and genetic evolution. They evolve a population of candidate solutions through mutation, crossover, and selection, gradually improving solutions over successive generations.

Each candidate solution, or \textit{individual}, is represented by a vector of parameters:

\[
r = \langle r_0, r_1, r_2, \dots, r_9 \rangle
\]

Each \( r_i \) corresponds to a physical acoustic parameter, detailed in Table~\ref{tab:params}. Parameters are constrained within realistic physical bounds to ensure feasible acoustic conditions; for instance, room dimensions are typically constrained between approximately \(2\,m\) and \(50\,m\).

\begin{table}
\centering
\caption{Parameters used in the acoustic configuration vector.}
\label{tab:params}
\begin{tabular}{cl}
\toprule
Index & Parameter \\
\midrule
\(r_0\) & Room length (\(m\)) \\
\(r_1\) & Room width (\(m\)) \\
\(r_2\) & Room height (\(m\)) \\
\(r_3\) & Microphone \(x\)-position (\(m\)) \\
\(r_4\) & Microphone \(y\)-position (\(m\)) \\
\(r_5\) & Microphone \(z\)-position (\(m\)) \\
\(r_6\) & Source \(x\)-position (\(m\)) \\
\(r_7\) & Source \(y\)-position (\(m\)) \\
\(r_8\) & Source \(z\)-position (\(m\)) \\
\(r_9\) & Surface absorption coefficient \\
\bottomrule
\end{tabular}
\end{table}

The fitness of each individual is computed by simulating the acoustic environment using PyRoomAcoustics, generating room impulse responses (RIRs), and applying these to reference speech audio samples. The fitness is evaluated based on two metrics: Word Error Rate (WER), measuring ASR transcription errors, and Short-Time Objective Intelligibility (STOI), a perceptual measure quantifying human speech intelligibility compared to the original audio. The goal is to identify configurations that increase ASR errors without significantly compromising human speech perception.

The evolutionary optimization proceeds as follows:

\begin{algorithm}[H]
\caption{Evolutionary Search for Acoustic Configuration}
\begin{algorithmic}[1]
\State Initialize population \(\mathcal{P}_0 = \{r^{(1)}, \dots, r^{(N)}\}\)
\For{generation \(g = 1\) to \(G\)}
    \For{each individual \(r^{(i)} \in \mathcal{P}_g\)}
        \State Simulate acoustic environment defined by \(r^{(i)}\)
        \State Generate impulse response and apply it to reference audio
        \State Compute WER and STOI for modified audio
        \State Calculate fitness from WER and STOI scores
    \EndFor
    \State Select best-performing individuals for mating pool
    \State Generate new population \(\mathcal{P}_{g+1}\) via crossover and mutation
\EndFor
\State \Return Best acoustic configuration found
\end{algorithmic}
\end{algorithm}

\subsubsection{Microphone Oscillation}

After optimizing the room configurations, we next enhance the robustness of our adversarial transformation through a simulated microphone oscillation technique.
The microphone oscillation technique simulates the effect of a continuously rotating microphone array, introducing dynamic directional variations into audio recordings. Our implementation achieves this by using multiple directional microphones, each positioned at a distinct fixed angle, capturing unique directional RIRs. 

To realistically emulate microphone rotation, we blend these directional RIRs using time-varying weighting functions. Specifically, for a given rotation frequency \(f_{rot}\), the blending weights \(w_i(t)\) for each directional microphone \(i\) are computed as raised-cosine (Hanning) windows. This ensures smooth transitions between impulse responses, resulting in perceptually natural audio dynamics:
\[
w_i(t) = 0.5\left(1+\cos\left(\frac{2\pi(t - t_i)}{T_{win}}\right)\right),
\]
where \(t_i\) is the temporal center for each microphone's window, and \(T_{win}\) is the window duration determined by the number of microphones. The weights are normalized so that their sum equals one at any given time, avoiding abrupt transitions.

The frequency of rotation \(f_{rot}\), measured in Hertz, dictates how quickly the microphone array completes a full rotation. For instance, at \(f_{rot}=1\,Hz\), a full rotation is simulated every second, closely replicating the scenario of a listener rotating their head or a microphone physically spinning.

This dynamic approach disrupts the temporal consistency relied upon by ASR models, challenging their robustness to directional acoustic variability, yet remaining comfortably intelligible to human listeners.

\subsubsection{Transient Acoustic Attenuation}

Lastly, we enhance EchoGuard using transient acoustic attenuation.

Transient acoustic attenuation refers to the brief, controlled reduction of audio amplitude within very short audio segments. Inspired by natural occurrences such as momentary dips in speech loudness during normal conversation, we introduce carefully timed attenuations, implemented using smooth Gaussian-shaped attenuation windows. This method selectively decreases amplitude in short segments (approximately 20-80 milliseconds), empirically observed to minimally affect human intelligibility while disproportionately degrading ASR performance.

Specifically, we define attenuation frames of length \( T_{frame} \), randomly selecting a subset \( p \) (typically 30\%) of frames to apply attenuation. Each selected frame undergoes amplitude modulation according to a Gaussian attenuation curve:
\[
    a(t) = 1 - (1 - \alpha) e^{-t^2},\quad t \in [-2,2]
\]
where \(\alpha\) is the attenuation factor controlling the depth of amplitude reduction. This approach maintains natural speech perception for humans while challenging ASR robustness.

\section{Evaluation}

First, we assess the effectiveness of our proposed jammer, \textit{EchoGuard}, in disrupting automatic speech recognition (ASR) systems. We benchmark EchoGuard against existing state-of-the-art (SOTA) jamming techniques under a black-box, zero-query, universal setting across a range of ASR models and datasets.

Second, we evaluate the \textit{practicality} of EchoGuard in real-world use. Specifically, we measure how well the jammer preserves human intelligibility and minimizes listener annoyance during live conversations. Since our defence is intended for deployment in potentially suspicious phone calls, many of which may still involve legitimate human callers, \textbf{a key requirement is that the jammer must not interfere with natural human-to-human communication}. We conduct a human study to quantify how pleasant and comprehensible the audio remains under different jamming methods, ensuring that our defence remains usable in everyday scenarios.

\subsection{Metrics}
\label{sec:metrics}

We evaluate audio quality, intelligibility, and jamming success rate using the following metrics:

\begin{itemize}
    \item \textbf{Word Error Rate (WER):} The WER measures the accuracy of automatic speech recognition (ASR) systems. It is defined as:

    \[
    \text{WER} = \frac{S + D + I}{W}
    \]

    where \(S\), \(D\), and \(I\) represent the number of substitutions, deletions, and insertions required to match the transcribed speech to the reference text, and \(W\) is the total number of words in the reference.

    \item \textbf{Short-Time Objective Intelligibility (STOI):} STOI is a quantitative measure of intelligibility for human listeners. It assesses the correlation between temporal envelopes of the original and processed speech signals over short time segments, resulting in a value between 0 and 1, where values closer to 1 indicate higher intelligibility \cite{taal2011algorithm}.

    \item \textbf{Jamming Success Rate:} The percentage of samples where the cosine similarity between clean and perturbed ASR transcriptions falls below 0.5. This threshold reflects the point at which perturbed transcripts no longer reliably convey the topic of the original message, based on user study findings~\cite{weiss2024your}. In LLM-driven vishing scenarios, even partial transcription failures can disrupt conversational coherence, making 0.5 a strong indicator of successful jamming.

    \item \textbf{Clarity:}: We asked users to rate clarity (reflecting how easy the speech was to understand) on a scale of 1-5, 1 (5 is best)

    \item \textbf{Pleasantness:} We asked users to rate pleasantness, reflecting subjective comfort and quality, on a scale of 1-5, (5 is best)
\item \textbf{Utility Index}. This metric captures the trade-off between ASR disruption and human perceptual quality. For each sample, we compute:
\[
\text{Utility} = \text{Pleasantness}_{\text{norm}} \times \text{Clarity}_{\text{norm}} \times \text{Degradation}_{\text{norm}}
\]
\begin{itemize}
    \item \textit{Pleasantness} and \textit{Clarity} scores are derived from 1--5 user ratings and normalized to the $[0, 1]$ range via $(x - 1)/4$.
    \item \textit{ASR Degradation} is calculated as $\frac{1 - \text{cosine similarity}}{2}$, mapping the cosine similarity range $[-1, 1]$ to $[0, 1]$. A value of 0 corresponds to no degradation (perfect similarity), while 1 indicates maximum degradation (complete dissimilarity).

\end{itemize}
The final Utility Index for a jamming method is the average Utility score across all evaluated samples.
\end{itemize}

\subsection{Datasets}
We evaluate \textsc{EchoGuard} using three publicly available datasets, chosen for their diverse speech content and wide adoption within ASR research:

\begin{itemize}
    \item \textbf{Tedlium (release 1)} consists of speech samples extracted from TED talks, providing diverse speakers and naturalistic speech patterns with varying speaking styles and speech rates \cite{tedlium}.
    
    \item \textbf{SPGISpeech} contains audio from corporate earnings calls, representing spontaneous and conversational speech in realistic business settings. The dataset captures natural dialogue variability and a range of acoustic environments, useful for real-world benchmarking \cite{panayotov2015librispeech}.
    
    \item \textbf{LibriSpeech} is derived from audiobooks and widely utilized for ASR benchmarks. To facilitate comparisons with existing results, our evaluations specifically employ the standard test and validation subsets \cite{librispeech}.
\end{itemize}

\subsection{ASR Jamming Baselines}
We benchmark \textsc{EchoGuard} against three prominent universal, real-time adversarial audio perturbation methods: \textbf{AdvDDoS}~\cite{ge2023advddos}, \textbf{Kenansville}~\cite{abdullah2021hear}, and \textbf{Kenku}~\cite{wu2023kenku}, as detailed in Section~\ref{sec:adv_attacks_asr}. These represent the types of black-box, zero-query attacks that a defender could realistically deploy within our novel defence framework~\textbf{ASRJam}. For \textbf{Kenansville}, we adopt the phoneme-level variant, which aligns with our real-time, zero-shot constraints and does not require per-sample optimization. Similarly, to maintain universal and real-time applicability with \textbf{Kenku}, we use a single pre-optimized perturbation associated with a generic command, avoiding input-specific customization.

To ensure fair and practical comparisons, we further adjusted the attack strengths across methods. Initial experiments indicated that Kenansville’s phoneme-level attacks, as originally implemented by the authors (targeting only a single phoneme), resulted in negligible impact on transcription accuracy. Consequently, we expanded this approach to attack all occurrences of the four most frequently appearing phonemes. Conversely, the original perturbations from Kenku and AdvDDoS proved excessively disruptive to listeners, significantly compromising audio quality. To balance adversarial effectiveness with perceptual practicality, we scaled down these perturbations by applying attenuation factors of \(0.2\) for AdvDDoS and \(0.33\) for Kenku, aligning the performance levels of all attacks to enable meaningful comparative evaluations.

\subsection{ASR models}

To evaluate the effectiveness of our proposed method, we benchmarked its performance against a diverse set of Automatic Speech Recognition (ASR) systems that an attacker might realistically use in a vishing pipeline. These include both open-source models and commercial APIs, representing a broad spectrum of commonly deployed ASR technologies. This setup allows us to assess how well our defence generalizes across the types of ASR systems likely to be leveraged by adversaries in real-world scenarios.

\begin{itemize}
    \item \textbf{DeepSpeech (DS)}: An open-source ASR model developed by Mozilla, known for its end-to-end deep learning architecture~\cite{hannun2014deep}. DeepSpeech utilizes a recurrent neural network to directly transcribe audio into text, offering a relatively simple and efficient approach. However, it can be sensitive to noise and variations in speech. We used a PyTorch implementation of DeepSpeech\footnote{\url{https://github.com/SeanNaren/deepspeech.pytorch}} in our evaluations.

    \item \textbf{Wav2Vec2 (W2V)}: A self-supervised learning framework for speech recognition developed by Facebook AI \cite{baevski2020wav2vec}. Wav2Vec2 learns powerful representations of speech from unlabelled data and fine-tunes them for specific ASR tasks, achieving state-of-the-art accuracy. While robust, it can be computationally intensive.
    \item \textbf{Vosk}: An open-source ASR toolkit that supports multiple languages \cite{Pantev2019VoskAS}. Vosk is designed to be lightweight and portable, suitable for use in various applications, including real-time systems. However, its accuracy may vary depending on the language and acoustic conditions.
    \item \textbf{Whisper}: A general-purpose speech recognition model developed by OpenAI, demonstrating strong performance across various languages and acoustic conditions \cite{radford2023robust}. Whisper is trained on a massive dataset of transcribed audio, exhibiting impressive robustness but with potential computational demands.
    \item \textbf{SpeechBrain (SB)}: An open-source toolkit for speech processing, including ASR \cite{Ravanelli2021SpeechBrain}. SpeechBrain provides a flexible and modular platform for developing and experimenting with different ASR models, offering a balance of accuracy and efficiency.
    \item \textbf{IBM Watson (IBM)}: A commercial cloud-based ASR service offered by IBM, providing robust speech-to-text capabilities optimized for enterprise applications \cite{ibmwatsonstt}. 

\end{itemize}

This combination of open-source and commercial ASR models allows for a comprehensive evaluation of our method's ability to disrupt a range of transcription technologies, reflecting real-world deployment scenarios where attackers may utilize various ASR services. 

% The diversity of these systems, in terms of their architecture, training data, and performance characteristics, strengthens the validity of our evaluation.

\subsection{EchoGuard Setup}

\textbf{Optimization and Configuration.} \textsc{EchoGuard} leverages an evolutionary algorithm to optimize acoustic parameters that influence reverberation and microphone oscillation characteristics. These parameters were constrained to physically plausible ranges and tuned for adversarial effectiveness. Table~\ref{tab:asrjam_parameters} summarizes the final optimized configuration. The genetic search was run with a population size of 200, crossover and mutation rates of 0.5, and executed for 11 generations. Fitness evaluation balanced ASR degradation, quantified by Word Error Rate (WER), with human intelligibility, measured by Short-Time Objective Intelligibility (STOI). Optimization was performed using a diverse subset of the Tedlium dataset to promote generalizability.

\begin{table}[htbp]
\centering
\caption{Optimized Parameters for \textsc{EchoGuard}}
\label{tab:asrjam_parameters}
\begin{tabular}{@{}lc@{}}
\toprule
\textbf{Parameter} & \textbf{Value} \\
\midrule
Room Length (m) & 68.55 \\
Room Width (m) & 58.89 \\
Room Height (m) & 20.73 \\
Mic Position (m) & (24.87, 23.59, 1.75) \\
Speaker Position (m) & (36.11, 0.97, 1.00) \\
Absorption Coefficient & 0.5 \\
\bottomrule
\end{tabular}
\end{table}

\textbf{EchoGuard Pipeline.} Once the room configuration was optimized, Room Impulse Responses (RIRs) were generated using PyRoomAcoustics. These RIRs were dynamically applied to each input signal using our directional microphone oscillation method. We set the rotation frequency to \(5\,\text{Hz}\), corresponding to five full microphone rotations per second. This simulated motion caused the audio signal to be modulated by a smoothly interpolated sequence of directional RIRs, introducing spatio-temporal distortions.

Additionally, transient acoustic attenuation was applied to the audio signal. The signal was divided into \(20\,\text{ms}\) frames, and \(30\%\) of frames were randomly selected for attenuation. Each affected segment was modulated using a Gaussian-shaped amplitude curve with a minimum scaling factor of \(0.7\), effectively creating subtle short-term disruptions aligned with high-energy regions of speech.

\begin{table*}
\centering
\caption{Jamming Success Rate across Datasets and ASR Systems (\textuparrow = higher is better)}
\label{tab:results_all}
\begin{tabular}{lcccccc}
\toprule
\textbf{Dataset / Method} & DS & W2V & Vosk & Whisper & SB & \cellcolor{gray!10}\textit{IBM} \\
\midrule

\multicolumn{7}{c}{\textbf{SPGISpeech}} \\
\midrule
Kenku & 0.9342 & 0.7227 & 0.5815 & 0.1350 & \textbf{0.6756} & \cellcolor{gray!10}\textit{0.6955} \\
Kenansville & 0.7508 & 0.4054 & 0.4486 & 0.0447 & 0.3157 & \cellcolor{gray!10}\textit{0.6688} \\
AdvDDoS & 0.9721 & 0.5600 & 0.4113 & 0.0953 & 0.6659 & \cellcolor{gray!10}\textit{0.4305} \\
Ours & \textbf{0.9918} & \textbf{0.8311} & \textbf{0.7639} & \textbf{0.1636} & 0.5965 & \cellcolor{gray!10}\textit{\textbf{0.7125}} \\

\midrule
\multicolumn{7}{c}{\textbf{LibriSpeech}} \\
\midrule
Kenku & 0.6696 & 0.1236 & 0.2999 & 0.0604 & 0.1025 & \cellcolor{gray!10}\textit{0.4748} \\
Kenansville & 0.2326 & 0.0470 & 0.1719 & 0.0522 & 0.0241 & \cellcolor{gray!10}\textit{0.3130} \\
AdvDDoS & 0.8578 & 0.1322 & 0.2907 & 0.0867 & \textbf{0.2130} & \cellcolor{gray!10}\textit{0.4611} \\
Ours & \textbf{0.9670} & \textbf{0.1489} & \textbf{0.4752} & \textbf{0.1419} & 0.0893 & \cellcolor{gray!10}\textit{\textbf{0.6333}} \\

\midrule
\multicolumn{7}{c}{\textbf{Tedlium}} \\
\midrule
Kenku & 0.7446 & 0.3004 & 0.2070 & 0.0383 & 0.2410 & \cellcolor{gray!10}\textit{0.4152} \\
Kenansville & 0.5388 & 0.1705 & 0.2433 & 0.0476 & 0.1603 & \cellcolor{gray!10}\textit{0.4558} \\
AdvDDoS & 0.9421 & 0.3202 & 0.2650 & 0.0786 & \textbf{0.4364} & \cellcolor{gray!10}\textit{0.4672} \\
Ours & \textbf{0.9886} & \textbf{0.4012} & \textbf{0.4596} & \textbf{0.1065} & 0.3123 & \cellcolor{gray!10}\textit{\textbf{0.6712}} \\

\bottomrule
\end{tabular}
\end{table*}

\subsection{Experiment Results}

\subsubsection{Jammer Success Rate}

To evaluate the effectiveness of \textsc{EchoGuard}, we report the \textit{jamming success rate} across three benchmark datasets: \textsc{SPGISpeech}, \textsc{LibriSpeech}, and \textsc{Tedlium}, against a diverse suite of ASR systems, including both open-source models and commercial APIs (Table~\ref{tab:results_all}).

Across the board, \textsc{EchoGuard} consistently outperforms all baseline jammers. Our method achieves the highest attack success rate on every ASR system tested, across all datasets, with only one minor exception: \textsc{SpeechBrain (SB)}, where it is slightly outperformed by the others. This anomaly is acceptable as \textsc{SpeechBrain} is relatively less common in real-world deployments as it's performance is poorer in general as an ASR system.

We also observe a general weakness of all jammers, including ours—against \textsc{OpenAI's Whisper}, which shows the lowest degradation among the ASR systems. However, \textsc{EchoGuard} still leads among all jammers on Whisper, outperforming the closest baseline by a notable margin. Importantly, while the absolute attack success rate on Whisper may seem modest (e.g., 0.14 on \textsc{LibriSpeech}), this still implies that 1 in 6 transcriptions is significantly corrupted, a degradation level that could be sufficient to disrupt scam conversations, especially in the context of interactive dialogue where misrecognition of key terms or intents can derail an LLM’s generation.

These results support our first key contribution: \textit{the viability of black-box, zero-query ASR jamming as a proactive defence}. While \textsc{EchoGuard} does not yet universally break all ASRs, particularly those like Whisper trained on massive, noisy, real-world data—it significantly raises the bar for adversarial robustness. In doing so, we not only highlight the vulnerability of real-time ASR in attack pipelines, but also push forward a new frontier in adversarial defence. We expect \textsc{EchoGuard} to inspire further development of adaptive ASR jammers and complementary countermeasures, reinforcing our broader \textsc{ASRJam} framework as a scalable, modular line of defence against AI-driven scams.

% To evaluate the effectiveness of EchoGuard, we compute the attack success rate across eight automatic speech recognition (ASR) systems... 

% % attack success rate was defined here before - I moved it up.
% % is defined as the percentage of samples where the cosine similarity between the transcriptions of clean and perturbed audio falls below 0.5, indicating a successful disruption of the original message. This threshold is based on findings from a user study which indicated that reconstructed sentences with a cosine similarity above 0.5 were generally perceived as revealing the topic of the original message \cite{weiss2024your}.  In the context of LLM-driven vishing attacks, disrupting even a few key details, rather than the entire subject, can be sufficient to confuse the LLM in a live, dynamic conversation where contextual accuracy is crucial. Therefore, our use of a 0.5 threshold represents a strong measure of attack success.

% Table~\ref{tab:results_all} report the results for SPGISpeech, LibriSpeech, and Tedlium respectively. \textsc{EchoGuard} outperforms the baseline methods across most open-source ASRs and achieves comparable or superior results on commercial APIs. When evaluating the APIs on the SPGISpeech dataset, we used a randomly selected 10\% subset (approximately 5 hours of audio)\footnote{The subset includes 500 segments randomly sampled from the official SPGISpeech test set, totaling roughly 5 hours.}.

\subsubsection{Practicality}
\label{sec:practicality}

While degrading ASR performance is essential, a critical dimension for the real-world deployment of our ASRJam framework is \textit{practicality}, encompassing both the effectiveness of the jamming and the subjective user experience. A viable jamming method must degrade transcription accuracy while preserving audio quality for human listeners, specifically regarding \textit{pleasantness} and \textit{clarity}. Methods that sound highly disruptive, uncomfortable, or unnatural may defeat the purpose of a human-compatible jammer and thus be impractical in everyday scenarios like phone calls.

To assess these practical dimensions, we conducted a human study. We selected 20 clean audio samples from the LibriSpeech dataset (10 from male speakers, 10 from female speakers). These 20 samples served as controls. We then applied our EchoGuard algorithm and the three baseline attack methods (Kenku, Kenansville, AdvDDoS) to each clean sample, creating 80 unique ``augmented'' samples. Participants in the study listened to all 100 samples (20 control + 80 augmented) presented in random order. For each sample, they provided ratings on a scale of 1 to 5 (where 5 is best) for both \textit{pleasantness} (reflecting subjective comfort and quality) and \textit{clarity} (reflecting how easy the speech was to understand).

\begin{figure}[t]
    \centering
    \includegraphics[width=\columnwidth]{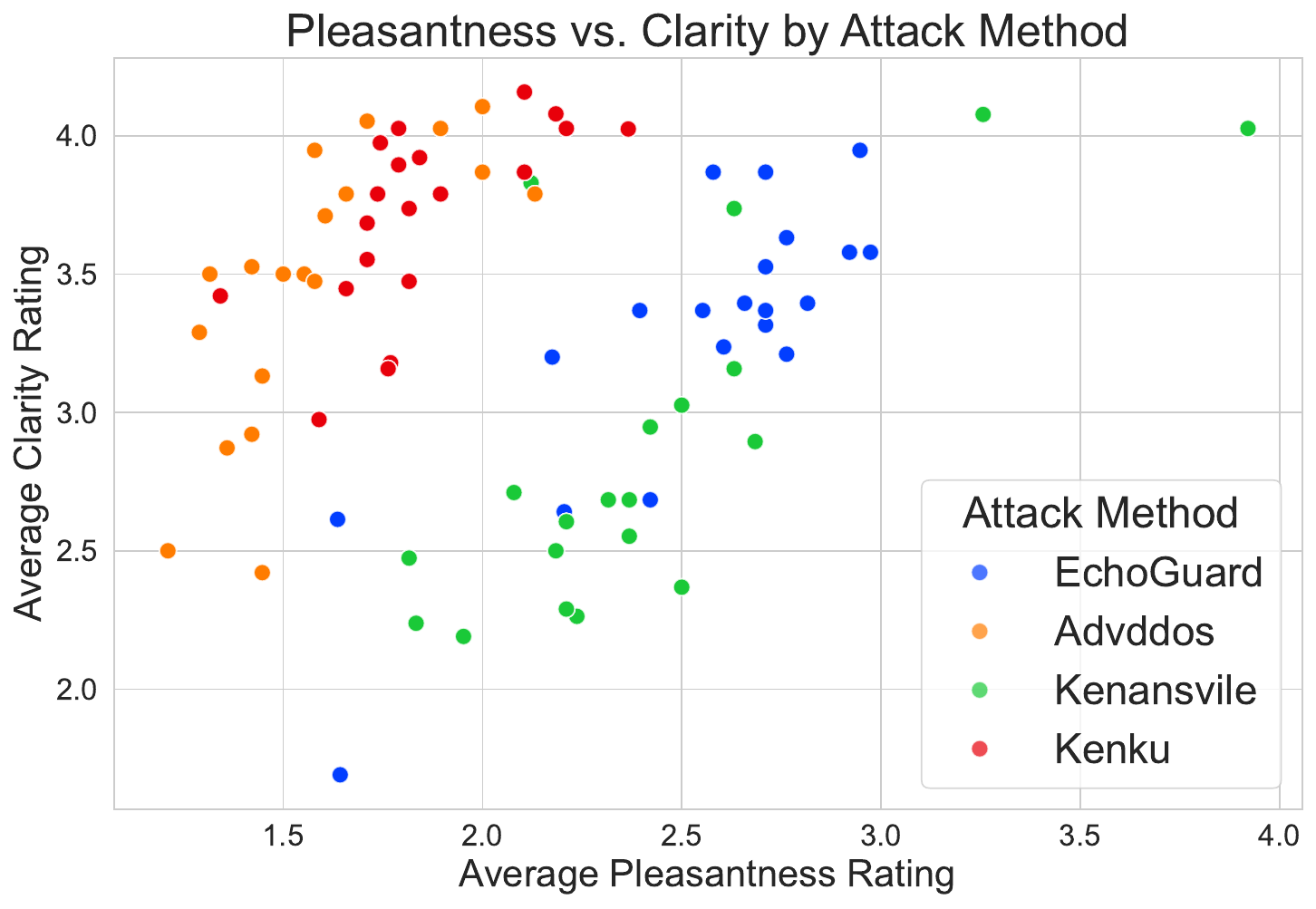}
    \caption{Clarity vs. Pleasantness for each sample and attack. Our method (EchoGuard) strikes the best balance between clarity and pleasantness.}
    \label{fig:scatter_practicality}
\end{figure}

Figure~\ref{fig:scatter_practicality} presents a scatter plot summarizing the user study results, showing average pleasantness versus average clarity for samples modified by each attack method. As observed, EchoGuard achieves significantly higher pleasantness ratings compared to the baseline methods. While Kenansville also yields some samples with reasonable pleasantness, participants often indicated difficulty comprehending speech under that attack, reflected in lower average clarity scores. Similarly while Kenku and Advddos had slightly higher clarity scores, their pleasantness was significantly lower. Therefore, EchoGuard thus appears to offer the best trade-off regarding overall user listening experience (balancing pleasantness and clarity).

User experience must be weighed against the primary goal: degrading ASR performance. To evaluate this trade-off comprehensively, we also processed the same 80 augmented samples (used in the user study) through various ASR models to assess their attack capability, measured by the cosine similarity between the original and perturbed transcriptions (lower similarity indicates higher degradation).To quantify this trade-off, we utilized the Utility Index defined in Section~\ref{sec:metrics}

\begin{figure*}[t]
    \centering
    \includegraphics[width=0.8\paperwidth]{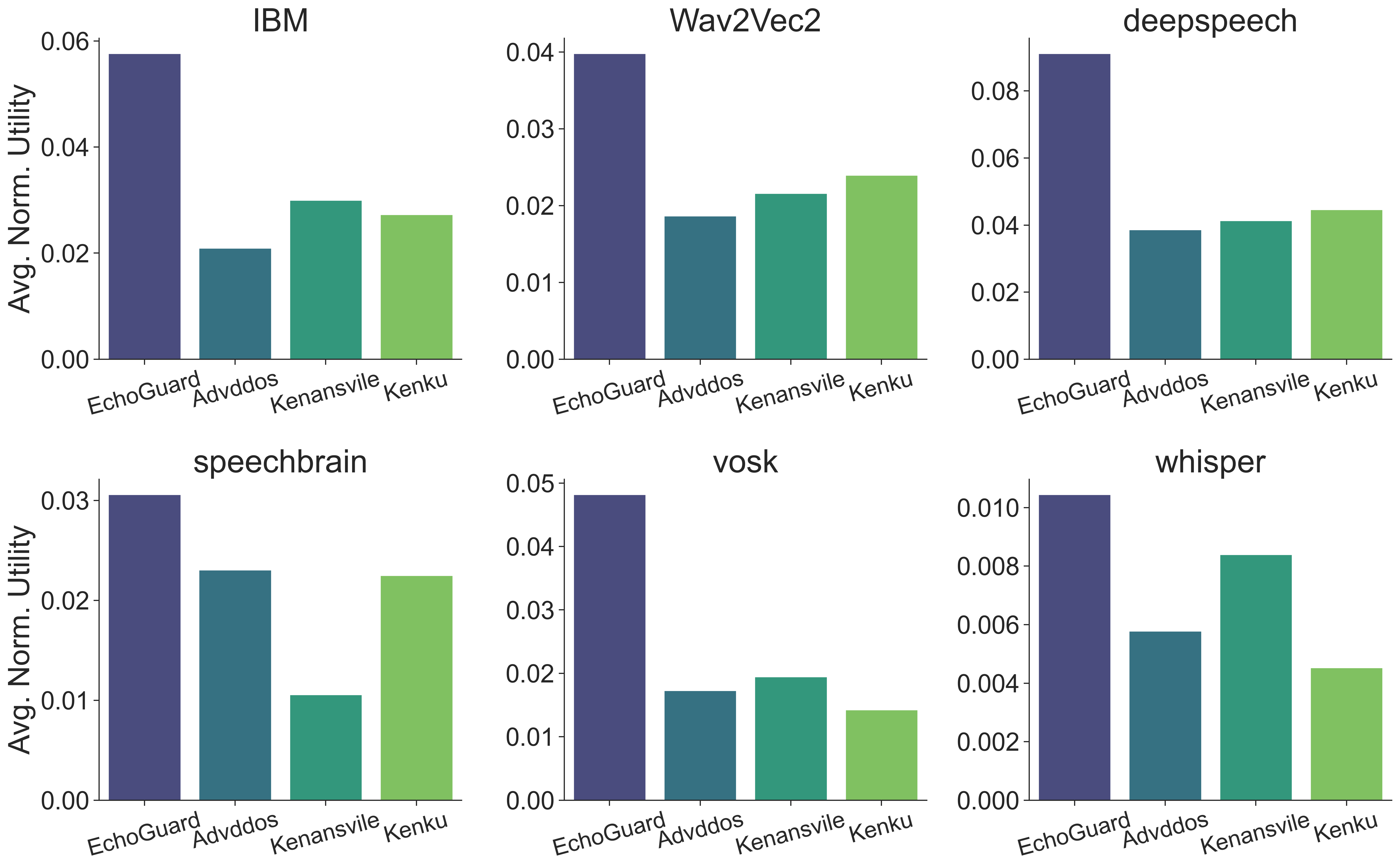}
    \caption{Average Utility Index across attacks. This metric quantifies the trade-off between transcription degradation and listener comfort.}
    \label{fig:utility_plot}
\end{figure*}

Figure~\ref{fig:utility_plot} shows the average Utility Index for EchoGuard and the baseline methods across the evaluated ASR models. The results clearly demonstrate that EchoGuard significantly outperforms the baselines, confirming its superior balance between practical user experience and adversarial effectiveness. These findings suggest that while other methods may achieve high clarity, their lower pleasantness or clarity limits their real-world applicability in interactive scenarios. EchoGuard, by design, successfully balances these competing goals, making it a more practical jamming solution for the proposed ASRJam defence framework.

\section{Discussion and Future Work}

The results presented in this work support the viability of using adversarial perturbations as a proactive defence mechanism against AI-driven voice scams. In particular, EchoGuard demonstrates that adversarial jamming techniques can be both effective in disrupting ASR pipelines and palatable to human users, thereby overcoming one of the most significant practical hurdles in real-world deployment.

A key insight from our experiments is that human-compatible distortions, derived from naturally occurring acoustic phenomena like reverberation, directional changes, and transient attenuation, offer a unique advantage. These transformations exploit a fundamental gap between human and machine auditory processing: while humans exhibit robust perceptual compensation for such distortions, ASR systems are brittle under similar conditions due to their dependence on clean and stable acoustic patterns. EchoGuard strategically amplifies this discrepancy.

From a deployment standpoint, our findings suggest that EchoGuard can serve as a drop-in, real-time defence for end-users who suspect an ongoing scam call. Importantly, the defence remains effective without requiring any knowledge of the adversary’s specific ASR model, making it broadly applicable in practical settings. The low perceptual impact observed in our user study reinforces its potential for everyday use, with minimal risk of degrading legitimate communication.

If an attacker attempted to fine-tune their ASR model on EchoGuard's perturbed samples, the resulting model would likely suffer degraded performance across clean and diverse real-world inputs due to overfitting to the perturbation \cite{wu2023kenku}. This makes large-scale adaptation costly and unattractive, further raising the bar for successful scam automation.

% Furthermore, while our utility index balances intelligibility, comfort, and effectiveness, actual deployment would require tuning for specific user preferences, threat environments, and device capabilities.

Finally, the utility index framework is easily tunable to favour stronger jamming when needed. For example, reducing the room's absorption coefficient increases reverberation, boosting ASR degradation at the expense of pleasantness, allowing users to adapt the defence to different threat levels or preferences.

\section{Conclusion}

In this paper, we introduced ASRJam, a proactive defence framework aimed at neutralizing AI-driven voice scams by targeting the speech recognition layer in the attacker’s pipeline. Our proposed jamming method, EchoGuard, exploits natural acoustic transformations that are cognitively tolerable for humans but detrimental to ASR systems. By leveraging zero-query, black-box, and universal perturbations, EchoGuard is uniquely positioned for real-time, model-agnostic deployment.

Through extensive evaluation across multiple ASR models and datasets, we demonstrated that EchoGuard consistently outperforms existing universal audio attacks in both adversarial effectiveness and human compatibility. It achieves significantly higher jamming success rates while preserving the clarity and pleasantness of the audio for legitimate callers. 

Ultimately, our work highlights a growing need for human-centered adversarial defences as generative AI systems become increasingly entangled with voice communication. EchoGuard and ASRJam offer a principled, scalable, and deployable line of defence against the next generation of automated social engineering threats.

\bibliographystyle{IEEEtran}
\bibliography{bib} 

\begin{thebibliography}{10}
\providecommand{\url}[1]{#1}
\csname url@samestyle\endcsname
\providecommand{\newblock}{\relax}
\providecommand{\bibinfo}[2]{#2}
\providecommand{\BIBentrySTDinterwordspacing}{\spaceskip=0pt\relax}
\providecommand{\BIBentryALTinterwordstretchfactor}{4}
\providecommand{\BIBentryALTinterwordspacing}{\spaceskip=\fontdimen2\font plus
\BIBentryALTinterwordstretchfactor\fontdimen3\font minus \fontdimen4\font\relax}
\providecommand{\BIBforeignlanguage}[2]{{%
\expandafter\ifx\csname l@#1\endcsname\relax
\typeout{** WARNING: IEEEtran.bst: No hyphenation pattern has been}%
\typeout{** loaded for the language `#1'. Using the pattern for}%
\typeout{** the default language instead.}%
\else
\language=\csname l@#1\endcsname
\fi
#2}}
\providecommand{\BIBdecl}{\relax}
\BIBdecl

\bibitem{openai2023gpt4}
OpenAI, ``Gpt-4 technical report,'' \emph{arXiv preprint arXiv:2303.08774}, 2023.

\bibitem{devlin2018bert}
J.~Devlin, M.-W. Chang, K.~Lee, and K.~Toutanova, ``Bert: Pre-training of deep bidirectional transformers for language understanding,'' in \emph{Proceedings of the 2019 Conference of the North American Chapter of the Association for Computational Linguistics: Human Language Technologies, Volume 1 (Long and Short Papers)}, 2019, pp. 4171--4186.

\bibitem{brown2020language}
T.~B. Brown, B.~Mann, N.~Ryder, M.~Subbiah, J.~Kaplan, P.~Dhariwal, A.~Neelakantan, P.~Shyam, G.~Sastry, A.~Askell \emph{et~al.}, ``Language models are few-shot learners,'' in \emph{Advances in neural information processing systems}, vol.~33, 2020, pp. 1877--1901.

\bibitem{huang2024gpt4phish}
Z.~Huang, L.~Xu, B.~Niu, Q.~Wang, and B.~Liang, ``Gpt-4-phish: Using gpt-4 to generate and evaluate phishing emails,'' \emph{arXiv preprint arXiv:2401.09727}, 2024.

\bibitem{li2024spearphishing}
K.~Li, J.~Zhang, K.~Chen, Y.~Wang, and S.~Jiang, ``Speargpt: Spear phishing emails generation via chatgpt,'' \emph{arXiv preprint arXiv:2406.13049}, 2024.

\bibitem{gressel2024exploiting}
G.~Gressel, R.~Pankajakshan, and Y.~Mirsky, ``Exploiting llms for scam automation: A looming threat,'' \emph{The 3rd ACM Workshop on the Security Implications of Deepfakes and Cheapfakes (WDC '24)}, 2024.

\bibitem{figueiredo2024viking}
J.~Figueiredo, T.~Gon{\c{c}}alves, D.~Ventura, H.~Gomes, D.~Correia, P.~H. Abreu, and M.~Monteiro, ``On the feasibility of fully ai-automated vishing attacks,'' \emph{arXiv preprint arXiv:2409.13793}, 2024.

\bibitem{toapanta2024aidriven}
J.~Toapanta, F.~Villamar{\'\i}n, and F.~Ord{\'o}{\~n}ez, ``Ai-driven vishing attacks: A practical approach,'' in \emph{Engineering Proceedings}, vol.~77, no.~1.\hskip 1em plus 0.5em minus 0.4em\relax MDPI, 2024, p.~15.

\bibitem{yu2024don}
Z.~Yu, X.~Liu, S.~Liang, Z.~Cameron, C.~Xiao, and N.~Zhang, ``Don't listen to me: understanding and exploring jailbreak prompts of large language models,'' in \emph{33rd USENIX Security Symposium (USENIX Security 24)}, 2024, pp. 4675--4692.

\bibitem{takemoto2024all}
K.~Takemoto, ``All in how you ask for it: Simple black-box method for jailbreak attacks,'' \emph{Applied Sciences}, vol.~14, no.~9, p. 3558, 2024.

\bibitem{muller2022does}
N.~M. M{\"u}ller, P.~Czempin, F.~Dieckmann, A.~Froghyar, and K.~B{\"o}ttinger, ``Does audio deepfake detection generalize?'' \emph{arXiv preprint arXiv:2203.16263}, 2022.

\bibitem{zhang2025audio}
B.~Zhang, H.~Cui, V.~Nguyen, and M.~Whitty, ``Audio deepfake detection: What has been achieved and what lies ahead,'' \emph{Sensors (Basel, Switzerland)}, vol.~25, no.~7, p. 1989, 2025.

\bibitem{carlini2018audio}
N.~Carlini and D.~Wagner, ``Audio adversarial examples: Targeted attacks on speech-to-text,'' in \emph{2018 IEEE security and privacy workshops (SPW)}.\hskip 1em plus 0.5em minus 0.4em\relax IEEE, 2018, pp. 1--7.

\bibitem{alzantot2018did}
M.~Alzantot, B.~Balaji, and M.~Srivastava, ``Did you hear that? adversarial examples against automatic speech recognition,'' \emph{arXiv preprint arXiv:1801.00554}, 2018.

\bibitem{neekhara2019universal}
P.~Neekhara, S.~Hussain, P.~Pandey, S.~Dubnov, J.~McAuley, and F.~Koushanfar, ``Universal adversarial perturbations for speech recognition systems,'' \emph{arXiv preprint arXiv:1905.03828}, 2019.

\bibitem{abdullah2021hear}
H.~Abdullah, M.~S. Rahman, W.~Garcia, K.~Warren, A.~S. Yadav, T.~Shrimpton, and P.~Traynor, ``Hear" no evil", see" kenansville": Efficient and transferable black-box attacks on speech recognition and voice identification systems,'' in \emph{2021 IEEE Symposium on Security and Privacy (SP)}.\hskip 1em plus 0.5em minus 0.4em\relax IEEE, 2021, pp. 712--729.

\bibitem{ge2023advddos}
Y.~Ge, L.~Zhao, Q.~Wang, Y.~Duan, and M.~Du, ``Advddos: Zero-query adversarial attacks against commercial speech recognition systems,'' \emph{IEEE Transactions on Information Forensics and Security}, 2023.

\bibitem{wu2023kenku}
X.~Wu, S.~Ma, C.~Shen, C.~Lin, Q.~Wang, Q.~Li, and Y.~Rao, ``$\{$KENKU$\}$: Towards efficient and stealthy black-box adversarial attacks against $\{$ASR$\}$ systems,'' in \emph{32nd USENIX Security Symposium (USENIX Security 23)}, 2023, pp. 247--264.

\bibitem{panayotov2015librispeech}
V.~Panayotov, G.~Chen, D.~Povey, and S.~Khudanpur, ``Librispeech: an asr corpus based on public domain audio books,'' in \emph{2015 IEEE international conference on acoustics, speech and signal processing (ICASSP)}.\hskip 1em plus 0.5em minus 0.4em\relax IEEE, 2015, pp. 5206--5210.

\bibitem{tedlium}
A.~Rousseau, P.~Del{\'e}glise, and Y.~Esteve, ``Ted-lium: an automatic speech recognition dedicated corpus.'' in \emph{LREC}, 2012, pp. 125--129.

\bibitem{librispeech}
V.~Panayotov, G.~Chen, D.~Povey, and S.~Khudanpur, ``Librispeech: an asr corpus based on public domain audio books,'' in \emph{2015 IEEE international conference on acoustics, speech and signal processing (ICASSP)}.\hskip 1em plus 0.5em minus 0.4em\relax IEEE, 2015, pp. 5206--5210.

\bibitem{hannun2014deep}
A.~Hannun, C.~Case, J.~Casper, B.~Catanzaro, G.~Diamos, E.~Elsen, R.~Prenger, S.~Satheesh, S.~Sengupta, A.~Coates, and A.~Y. Ng, ``Deep speech: Scaling up end-to-end speech recognition,'' \emph{arXiv preprint arXiv:1412.5567}, 2014.

\bibitem{baevski2020wav2vec}
A.~Baevski, H.~Zhou, A.~Narayanan, W.-H. Senior, and M.~Hughes, ``wav2vec 2.0: A framework for self-supervised learning of speech representations,'' \emph{arXiv preprint arXiv:2006.11477}, 2020.

\bibitem{Pantev2019VoskAS}
A.~Pantev, S.~Nikolay, P.~Petr, S.~Alexander, and K.~Evgenii, ``Vosk: An open source offline speech recognition toolkit,'' in \emph{Proceedings of the 20th Conference of the European Chapter of the Association for Computational Linguistics: Companion Volume Proceedings of the Student Research Workshop}, 2019, pp. 21--28.

\bibitem{radford2023robust}
A.~Radford, J.~W. Kim, T.~Xu, G.~Brockman, C.~McLeavey, and I.~Sutskever, ``Robust speech recognition via large-scale weak supervision,'' in \emph{International conference on machine learning}.\hskip 1em plus 0.5em minus 0.4em\relax PMLR, 2023, pp. 28\,492--28\,518.

\bibitem{Ravanelli2021SpeechBrain}
M.~Ravanelli, J.~Parcollet, P.~Plantinga, A.~Rouhe, T.~Cornell, L.~Lugosch, M.~Matassoni, C.~Subakan, J.~Turrin, S.~Watanabe \emph{et~al.}, ``Speechbrain: A general-purpose speech toolkit,'' in \emph{2021 IEEE Automatic Speech Recognition and Understanding Workshop (ASRU)}.\hskip 1em plus 0.5em minus 0.4em\relax IEEE, 2021, pp. 222--229.

\bibitem{prabhavalkar2023end}
R.~Prabhavalkar, T.~Hori, T.~N. Sainath, R.~Schlüter, and S.~Watanabe, ``End-to-end speech recognition: A survey,'' \emph{IEEE/ACM Transactions on Audio, Speech, and Language Processing}, vol.~31, pp. 1--17, 2023.

\bibitem{kheddar2024automatic}
H.~Kheddar, M.~Hemis, and Y.~Himeur, ``Automatic speech recognition using advanced deep learning approaches: A survey,'' \emph{Information Fusion}, p. 102422, 2024.

\bibitem{yuan2018commandersong}
X.~Yuan, Y.~Chen, Y.~Zhao, Y.~Long, X.~Liu, K.~Chen, S.~Zhang, H.~Huang, X.~Wang, and C.~A. Gunter, ``$\{$CommanderSong$\}$: A systematic approach for practical adversarial voice recognition,'' in \emph{27th USENIX security symposium (USENIX security 18)}, 2018, pp. 49--64.

\bibitem{schonherr2018adversarial}
L.~Sch{\"o}nherr, K.~Kohls, S.~Zeiler, T.~Holz, and D.~Kolossa, ``Adversarial attacks against automatic speech recognition systems via psychoacoustic hiding,'' \emph{arXiv preprint arXiv:1808.05665}, 2018.

\bibitem{taori2019targeted}
R.~Taori, A.~Kamsetty, B.~Chu, and N.~Vemuri, ``Targeted adversarial examples for black box audio systems,'' in \emph{IEEE Deep Learning and Security Workshop}, 2019, arXiv:1805.07820.

\bibitem{liu2024difattack}
J.~Liu, J.~Zhou, J.~Zeng, and J.~Tian, ``Difattack: Query-efficient black-box adversarial attack via disentangled feature space,'' in \emph{Proceedings of the AAAI Conference on Artificial Intelligence}, vol.~38, no.~4, 2024, pp. 3666--3674.

\bibitem{ambra}
\BIBentryALTinterwordspacing
A.~Demontis, M.~Melis, M.~Pintor, M.~Jagielski, B.~Biggio, A.~Oprea, C.~Nita-Rotaru, and F.~Roli, ``Why do adversarial attacks transfer? explaining transferability of evasion and poisoning attacks,'' in \emph{28th USENIX Security Symposium (USENIX Security 19)}.\hskip 1em plus 0.5em minus 0.4em\relax Santa Clara, CA: USENIX Association, Aug. 2019, pp. 321--338. [Online]. Available: \url{https://www.usenix.org/conference/usenixsecurity19/presentation/demontis}
\BIBentrySTDinterwordspacing

\bibitem{wang2024practicalsurveyemergingthreats}
\BIBentryALTinterwordspacing
Y.~Wang, Q.~Yan, N.~Ivanov, and X.~Chen, ``A practical survey on emerging threats from ai-driven voice attacks: How vulnerable are commercial voice control systems?'' 2024. [Online]. Available: \url{https://arxiv.org/abs/2312.06010}
\BIBentrySTDinterwordspacing

\bibitem{chen2020devil}
Y.~Chen, X.~Yuan, J.~Zhang, Y.~Zhao, S.~Zhang, K.~Chen, and X.~Wang, ``$\{$Devil’s$\}$ whisper: A general approach for physical adversarial attacks against commercial black-box speech recognition devices,'' in \emph{29th USENIX Security Symposium (USENIX Security 20)}, 2020, pp. 2667--2684.

\bibitem{tsironis2024adaptation}
A.~Tsironis, E.~Vlahou, P.~Kontou, P.~Bagos, and N.~Kop{\v{c}}o, ``Adaptation to reverberation for speech perception: A systematic review,'' \emph{Trends in Hearing}, vol.~28, p. 23312165241273399, 2024.

\bibitem{luo2024cortical}
C.~Luo and N.~Ding, ``Cortical encoding of hierarchical linguistic information when syllabic rhythms are obscured by echoes,'' \emph{NeuroImage}, vol. 300, p. 120875, 2024.

\bibitem{garcia2024sensory}
H.~G. Garc{\'\i}a-L{\'a}zaro and S.~Teng, ``Sensory and perceptual decisional processes underlying the perception of reverberant auditory environments,'' \emph{eneuro}, vol.~11, no.~8, 2024.

\bibitem{gao2024original}
J.~Gao, H.~Chen, M.~Fang, and N.~Ding, ``Original speech and its echo are segregated and separately processed in the human brain,'' \emph{PLoS Biology}, vol.~22, no.~2, p. e3002498, 2024.

\bibitem{taal2011algorithm}
C.~H. Taal, R.~C. Hendriks, R.~Heusdens, and J.~Jensen, ``An algorithm for intelligibility prediction of time--frequency weighted noisy speech,'' \emph{IEEE Transactions on audio, speech, and language processing}, vol.~19, no.~7, pp. 2125--2136, 2011.

\bibitem{weiss2024your}
R.~Weiss, D.~Ayzenshteyn, and Y.~Mirsky, ``What was your prompt? a remote keylogging attack on $\{$AI$\}$ assistants,'' in \emph{33rd USENIX Security Symposium (USENIX Security 24)}, 2024, pp. 3367--3384.

\bibitem{ibmwatsonstt}
\BIBentryALTinterwordspacing
\emph{IBM Watson Speech to Text}, IBM, 2025. [Online]. Available: \url{https://www.ibm.com/products/speech-to-text}
\BIBentrySTDinterwordspacing

\end{thebibliography}

\end{document}